# Cable Driven Rehabilitation Robots: Comparison of Applications and Control Strategies

**MUHAMMAD SHOAIB**[1] **(Member, IEEE), EHSAN ASADI**[1]**(Senior Member, IEEE), JOONO CHEONG**[2]**(Member, IEEE) and ALIREZA BAB-HADIASHAR**[1]**(Senior Member, IEEE)**
[1]School of Engineering, RMIT University, Melbourne, VIC 3000, Australia
[2]Department of Electro-Mechanical System Engineering, Korea University, Sejoing-City, Korea.

Corresponding author: Muhammad Shoaib (muhammad.shoaib@rmit.edu.au).

**ABSTRACT** Significant attention has been paid to robotic rehabilitation using various types of actuator and power transmission. Amongst those, cable-driven rehabilitation robots (CDRRs) are relatively newer and their control strategies have been evolving in recent years. CDRRs offer several promising features, such as low inertia, lightweight, high payload-to-weight ratio, large work-space and configurability. In this paper, we categorize and review the cable-driven rehabilitation robots in three main groups concerning their applications for upper limb, lower limb, and waist rehabilitation. For each group, target movements are identified, and promising designs of CDRRs are analyzed in terms of types of actuators, controllers and their interactions with humans. Particular attention has been given to robots with verified clinical performance in actual rehabilitation settings. A large part of this paper is dedicated to comparing the control strategies and techniques of CDRRs under five main categories of: Impedance-based, PID-based, Admittance-based, Assist-as-needed (AAN) and Adaptive controllers. We have carefully contrasted the advantages and disadvantages of those methods with the aim of assisting the design of future CDRRs.

## I. INTRODUCTION

ACCORDING to a report published by the World Health Organization [1], there are over a billion of individuals, almost 15% of the world's population, facing some kind of movement-related disability, mainly because of stroke, intellectual impairment, traumatic brain injury, musculoskeletal and neurological disorders [2]. Also, movement-related disabilities are often seen in older people. There are strong evidence that these movement-related disabilities can be restored by intensive repetitive rehabilitation training [3], [4].

Physiotherapy is a common approach for rehabilitation, which has limitations in terms of availability and intensity of therapists. Additionally, manual training is time consuming, expensive and also depends on therapist's experience. Robotic rehabilitation offers promising features including repetitive training with uniform performance for a long period of time and quantitative measures for performance analysis. It can also reduce labour intensity as well as cost, and improve the efficiency of rehabilitation process.

Typically, rehabilitation robots are classified into three main categories, end-effector, exoskeleton, and planar based on their kind of attachment to the user [5]. In an end-effector type [6], [7], the end-effector of robot is generally attached to the user body while in an exoskeleton type, also known as wearable rehabilitation robots, the robot is wrapped around the human body, and each joint is controlled independently [8]–[10]. In the planar type, robot only allows movements of its attachment points to the body in a specific plan [11].

The common actuation techniques in rehabilitation robots include, pneumatic, hydraulic, series elastic and electric. Hydraulic and pneumatic actuators are bulky, noisy and has possibility of compressed air or fluid leakage, which may limit their usage. For power transmissions, different approaches are used including ball screw driven, belt driven, gear driven, and cable driven [12]. Important characteristics of rehabilitation robots, in terms of power transmissions, are compared in Table 1, where the desirable characteristics are underlined [13]–[16]. Compared to other types, CDRRs offer several promising features such as: low inertia, light weight, high payload-to-weight ratio, and large workspace [17]–[20]. They also have a few deficiencies due to unidirectional power transmission, vibration and maintenance.

Although there are a large number of review papers on rehabilitation robots [21]–[43], only one has so far focused on cable driven rehabilitation robots [44]. The paper analyzes CDRRs in terms of safety, back-drivability, weight, adaptability, versatility, and misalignment features. The paper however does not include other important aspects such as control strategies, clinical studies, physical human-robot interaction and human-machine interfaces, which we have covered in this work to complete the analysis. The control strategies are critical and different among various transmission approaches. Due to the elasticity of cables and the fact they can only apply tensile forces [45], [46], many of the control strategies deployed in robots with gears, belts, ball-screws, linkages, or others [25], [27], [31], [34], [43], [47], cannot be directly applied to CDRRs [18], [20].

In this paper, we categorize the CDRRs into three main groups concerning their application for upper limb, lower limb, and waist rehabilitation, as shown in Fig. 1. For each group, target movements are identified, and existing CDRRs are reviewed in terms of types of actuators, controllers, clinical studies, physical human-robot interactions, and human-machine interfaces (these are complimentary to the topics presented in [44]). Particular attention has been given to clinical performance and results of CDRR's in actual rehabilitation environment. A large part of this work is dedicated to compare the control strategies and techniques used for CDRRs. These are divided into five main categories including Impedance-based, PID-based, Admittance-based, Assist-as-needed, and Adaptive-based controllers. We conclude the paper by identifying major challenges facing the future development of CDRRs and their suitability for rehabilitation.

The classification schema is explained in section II. Section III reviews the design characteristics of existing CDRRs for upper, lower and waist rehabilitations. Then the control strategies of CDRRs are presented in section IV. Lastly, in section V challenges and future works are summarized.

TABLE 1: Comparison of cable, belt, ball screw and gear driven rehabilitation robots

| Properties | CDRR | BDRR | BSDRR | GDRR |
|---|---|---|---|---|
| Distance between source and load/Workspace | Large | Large | Medium | Small |
| Manipulation accuracy at low speed | High | Medium | Medium | High |
| Payload-to-weight ratio | High | High | Low | Low |
| Moving inertia of robot's link | Low | Medium | High | High |
| Weight of rehabilitation robot | Low | Low | High | High |
| Maintenance | High | Medium | Low | Low |
| Direction of power transmission | Uni-Dir | Bi-Dir | Bi-Dir | Bi-Dir |
| Undesired vibration | High | High | Low | Low |

## II. PRELIMINARIES AND CLASSIFICATION OF CABLE-DRIVEN ROBOTS FOR REHABILITATION

Conventionally, rehabilitation robots are classified into only two groups of upper and lower limbs as there are significant differences in motor function, mobility and gait of the upper and lower extremity of human body. The application of rehabilitation robots has been extended in recent years to other parts of the human body (e.g. waist) and the related aches caused by diseases, old age, or sedentary lifestyle. The differences in motor function and mobility of upper limb, lower limb and waist also led researchers to design and implement specific CDRRs for different extremities of human body. In this work, the existing cable driven rehabilitation robots are preliminarily classified into three groups based on their applications in upper limb, lower limb, and waist rehabilitation. Around 200 references and research articles, which present a novel or enhanced cable-driven rehabilitation mechanism, or discussing key technologies, design characteristics or control strategies for CDRRs, have been reviewed in this work. Among those references, 86 are recent works published after 2015. Overall, we have identified and summarized 66, 31 and 4 state-of-the-art CDRRs for upper limb, lower limb and waist rehabilitation, respectively. To facilitate the discussion, the CDRRs belonging to each group (upper limb, lower limb and waist) are then classified and compared based on the design characteristics and control strategies, as shown in Fig. 1. The classification and review schema is as follows:

*Classification and review schema*
I. Rehabilitation application:
- Upper limb: elbow, shoulder, wrist, forearm, fingers.
- Lower limb: hip, knee, ankle and foot.
- Waist rehabilitation.

II. Design characteristics:
- Target joint movements: single, multiple or all joints.
- Types of actuators: pneumatic, hydraulic, series elastic, and electric actuators.
- Type of sensors: IMU, force, torque, EMG, load cell.
- Human-robot physical interaction: exoskeleton, end-effector, and planar type.
- Human-machine interface: virtual, visual, or tactile
- Clinical study: levels 1, 2, 3, and 4.

III. Control strategies (impedance, PID, admittance, AAN, and adaptive controllers):
- Control's level: high, mid, and low-level.
- Modelling approach: model-based, model-free, and hybrid control.
- Performance: measurements of error, cartesian error for task space tracking, and joints trajectory tracking error.
- Operating space: task and configuration space.
- Stability: asymptotic, non-asymptotic and not applicable.

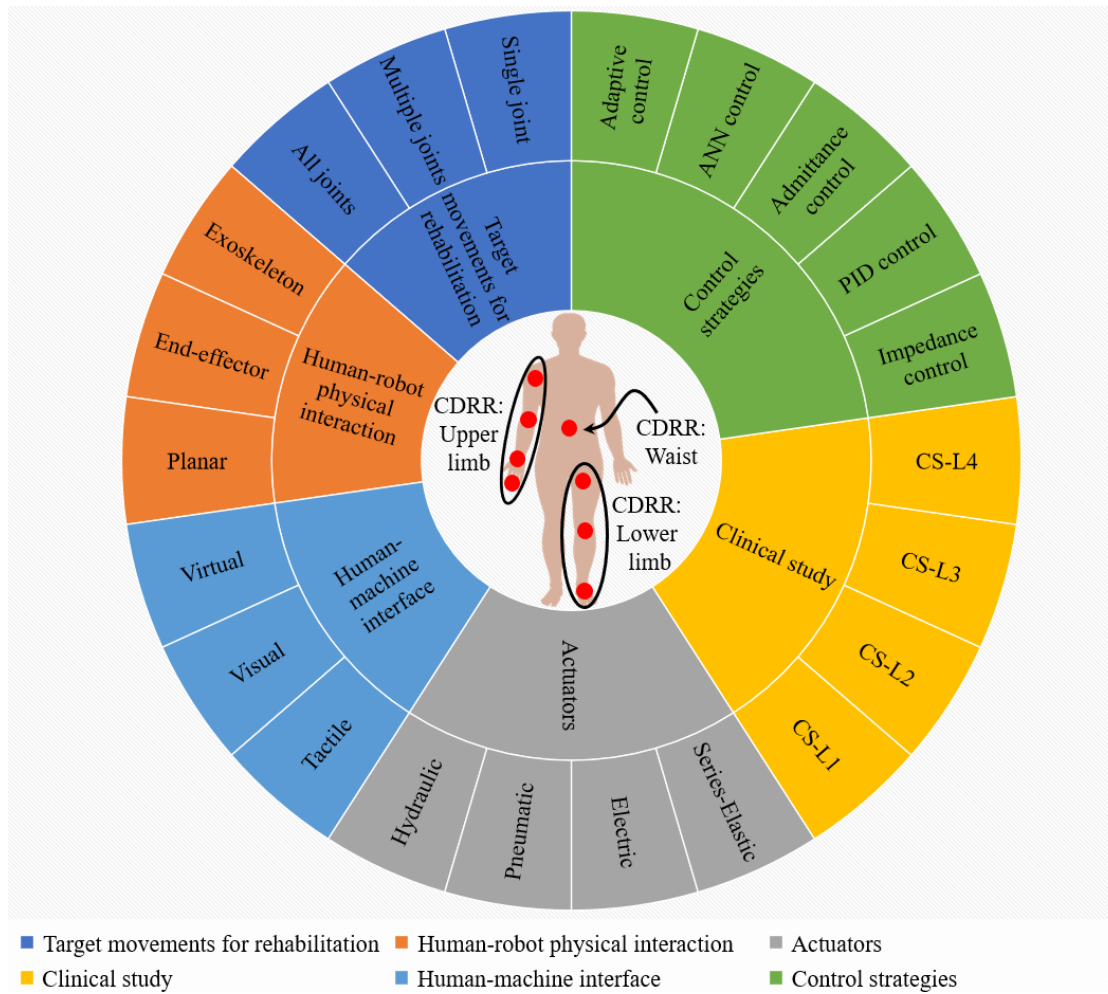

FIGURE 1: The review structure and classification schema considering application, design characteristics and control strategies. Around 66, 31 and 4 different CDRRs are found for upper limb, lower limb and waist rehabilitation, respectively.

### A. TARGET JOINT MOVEMENTS

Target movements refer to the specific movements in human joints for which the rehabilitation robot is designed. Based on the target movements, CDRRs can be divided into three categories; CDRRs that can assist and train only single joint, multiple joints, and all joints of upper or lower limbs.

### B. TYPE OF ACTUATORS

Different types of actuators are used in the cable driven rehabilitation robots, which are electric, hydraulic, series elastic, and pneumatic. An actuator in CDRRs is responsible for generation of mechanical energy and motion [48].

*Electric actuators* convert electric energy into mechanical energy. There are large varieties of commercially available electric actuators with different specifications used in CDRRs. *Pneumatic* and *Hydraulic actuators* convert highly compressed air and hydraulic fluid pressure into mechanical energy, respectively. *Series Elastic Actuators (SEA)* in CDRRs use an elastic element between the actuator and load.

### C. HUMAN-ROBOT PHYSICAL INTERACTION (HRPI)

The human-robot physical interaction points to how the rehabilitation robots are attached to the patients. We categorize the CDRRs into three groups; end-effector, exoskeleton, and planner type CDRRs based on human-robot physical interaction. The used terminologies are explained in the Table 2. The end-effector, exoskeleton, and planner type CDRRs can be further categorized into serial and parallel mechanisms. *Exoskeleton type CDRRs*, also known as wearable robots, have mechanical structure correspondence to the actuated skeleton structure. An *End-effector type CDRR*, has its end-effector in contact with only one segment of the human body or upper limb only at its most distal part. A *planar type CDRR*, provides movements in a specific plane but allows limb to move in three-dimensional space.

### D. HUMAN-MACHINE INTERFACE (HMI)

Human-Machine Interface refers to a type of virtual, visual, or tactile interface that is usually incorporated in rehabilitation exercises to allow patient or user interaction with robot, HMI could also increase the recovery rate of rehabilitation.

TABLE 2: Glossary: human-robot physical interaction

| Term | Description |
| --- | --- |
| E-CDRRs: Exoskeleton Type | CDRRs resemble the skeletal structure of the human limb and connect to patients at multiple points. Robot's joint axes have one to one correspondence with the human joints. |
| EE-CDRRs: End-effector Type | CDRRs are connected to patients only at the most distal part of the extremity. |
| PL-CDRRs: Planar Type | CDRRs allow movements of end-effector in a specific plane. Even though the robot's movements is performed in a single plane, but limb joints may move in a three-dimensional space. |

### E. CLINICAL STUDY

The clinical study refers to the verification of rehabilitation robots in practice with real patients. It is an essential step in evaluating the actual performance of CDRRs, control strategies, and various aspects of robot interaction with patients. However, conducting clinical trials is expensive, challenging, and needs a higher level of reliability and safety measures to verify the robot in actual settings with multiple humans or patients involved. While clinical trials are of great importance, there is less information on clinical studies for different CDRRs and those might depend on unique cases and cable configurations. Particular attention is given in this work to summarize the major information on existing clinical trials for CDRRs with the aim of assisting the design and assessments of future CDRRs through clinical studies.

The clinical trials with patients and/or healthy human are compared here in four levels as outlined in Table 3. *Initial clinical study or level 1* is carried out with only few healthy volunteers, to evaluate a CDRR. *Level 2* clinical study is carried out with large healthy volunteers or less than 30 individuals suffering from the target disease, to check performance in assistance and training the targeted disability. *Level 3* clinical study is performed with 30 to 100 patients suffering from target disease. *Level 4* is the final level of clinical trial with more than 100 patients suffering from a target disease for further verification of the CDRR performance.

## III. DESIGN CHARACTERISTIC OF CDRRS

Main features or design characteristics of the existing CDRRs for upper limb, lower limb and waist rehabilitation are discussed in separate subsections. Approximately 66, 31 and 4 different designs of CDRRs are found in literature for upper limb, lower limb, and waist rehabilitation, respectively. The advantages and disadvantages of different characteristics are summarised in the last subsection.

### A. CABLE-DRIVEN ROBOTS FOR UPPER LIMB

There are large number of CDRRs designed for patients with upper limb disabilities. A few key technologies in cable-driven robots for upper limb rehabilitation are CAREX [8], [49], Dampace [50], Planar Cable-Driven Robot [51], [52], MEDARM [11], PACER [53] [54], and Active therapeutic device (ATD) [55] as shown in Fig. 2. A complete list of existing CDRRs for upper limb rehabilitation are summarized

TABLE 3: Glossary of terms: classification of clinical study

| Term | Description |
| --- | --- |
| Level 1 CS-L1 | Initial study is carried out with only few healthy volunteers, to check the CDRR efficiency, safety, and controllability. It also involves evaluating the CDRR performance with external disturbance. |
| Level 2 CS-L2 | Trials are performed with large healthy volunteers or less than 30 individuals suffering from the target disease, to evaluate the CDRR performance, effectiveness in assistance, and training the targeted disability. |
| Level 3 CS-L3 | Trials are performed with 30 to 100 patients suffering from the target disease. It is conducted to verify the ability in assist and training the targeted disability. |
| Level 4 CS-L4 | Final level of clinical study is performed with a large number of patients suffering from the target disease for further verification of the CDRR performance. |

in Tables 4, 5, and 6.

**Target movements for rehabilitation:** Based on the target movements, CDRRs for upper limb can be divided into three categories. The first category includes CDRRs that can assist and train only one joint movement, which can be shoulder, elbow, forearm, wrist or finger movement, as summarized in Table 4. A few examples are, a) MEDARM [56], [57], Wearable Soft Orthotic Device [58], HRM [59], and Active Soft Orthotic Device [60] to assist shoulder; b) NEUROExos [61], CADEL [62] and BJS [63] to assist elbow; c) Active Therapeutic Device (ATD) [55] to assist forearm; d) CDWRR [64] to assist wrist; and e) Exoskeleton glove [65], Tendon Driven Hand orthosis [66], CDRH [67], Home-based hand rehabilitation [68], wearable robot hand [69] and IOTA [70] to assist fingers joints. The second category of CDRRs, shown in Table 5, is designed to assist a combination of two or more joints movements of the upper limb, for example, ULERD [71], SUEFUL-7 [72], CABXLexo-7 [73] and CAREX-7 [9], [74]. The third category of CDRRs, as summarized in Table 6, can assist and train all the joints of upper limb except fingers. There doesn't seem to be any CDRR that can train the whole arm including fingers.

**Types of actuators:** Electric actuators are widely used in the upper limb CDRRs, such as ULERD [71], [75], SUEFUL-7 [72], CAREX [8], CDWRR [64]. Compare to electric actuators, the use of pneumatic actuators in CDRRs has gained less attention from the research community. IKO [76] [77], Exoskeleton Rehabilitation Robot [78] , 9-DoFs rehabilitation robot [79], [80], RUPERT [81] [82] [83], Wearable Rehabilitation Robotic Hand [84] are a few CDRRs that have been designed using pneumatic actuators. There are currently only two upper limb CDRRs, NEUROExos [61], Dampace [50], which use Hydraulic actuators. Series Elastic Actuators are also found in four upper limb CDRRs, namely: Wearable Soft Orthotic Device [58], Active Soft Orthotic Device [60], BJS [63], and Elbow Exoskeleton [85]. While each type of actuator has some advantages and disadvantages, one often selects the best choice for a specific rehabilitation application based on the most critical features that are nonnegotiable.

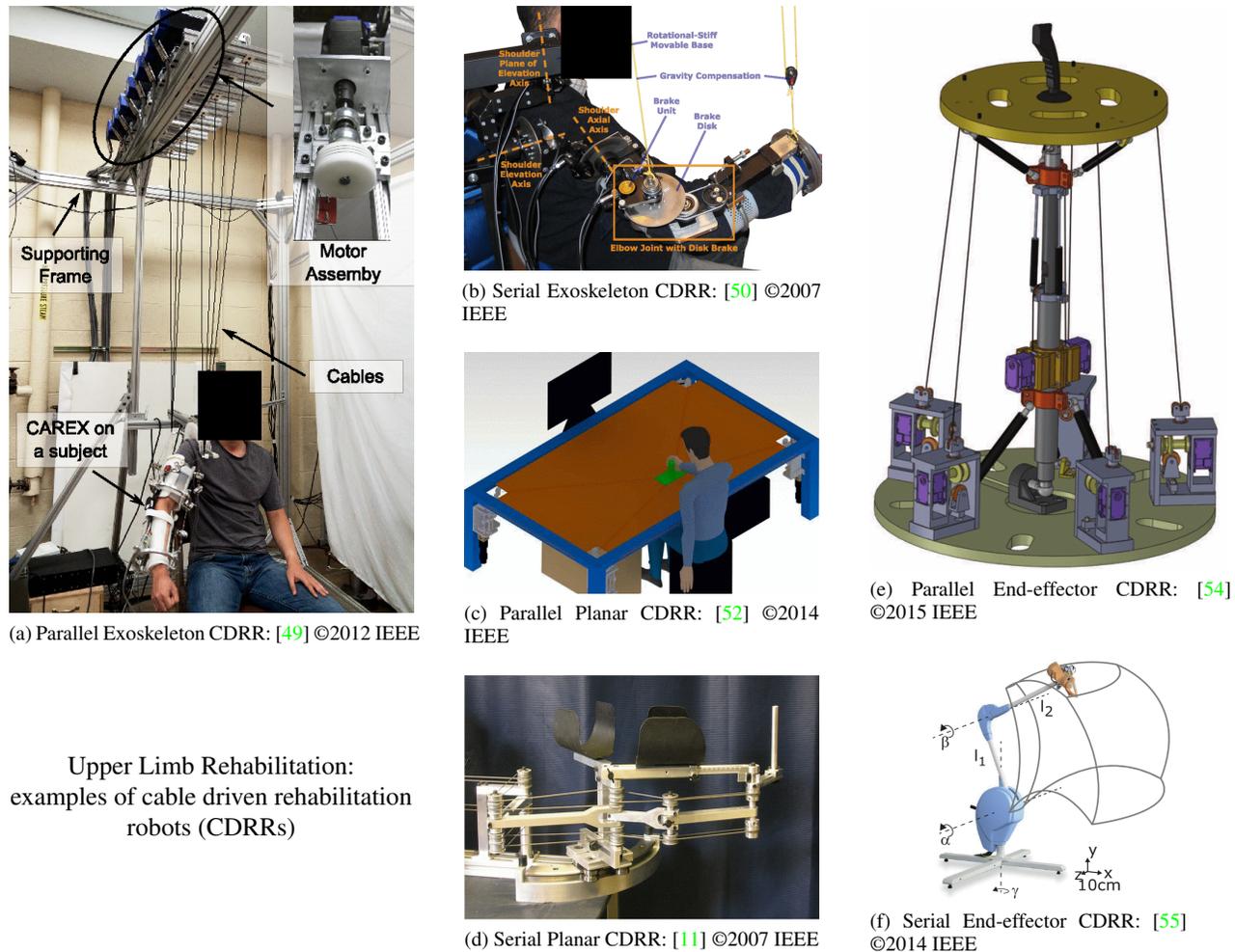

(a) Parallel Exoskeleton CDRR: [49] ©2012 IEEE

(b) Serial Exoskeleton CDRR: [50] ©2007 IEEE

(c) Parallel Planar CDRR: [52] ©2014 IEEE

(d) Serial Planar CDRR: [11] ©2007 IEEE

(e) Parallel End-effector CDRR: [54] ©2015 IEEE

(f) Serial End-effector CDRR: [55] ©2014 IEEE

Upper Limb Rehabilitation: examples of cable driven rehabilitation robots (CDRRs)

FIGURE 2: Upper limb rehabilitation: examples of cable driven robots (CDRRs). **(a)**: Parallel Exoskeleton type CDRR - CAREX [49] ©2012 IEEE. E/Reprinted, with permission from "Transition from mechanical arm to human arm with CAREX: A cable driven ARm EXoskeleton (CAREX) for neural rehabilitation" by Ying Mao et al. in IEEE International Conference on Robotics and Automation 2012; **(b)**: Serial Exoskeleton type CDRR - Dampace [50] ©2007 IEEE. E/Reprinted, with permission from "Dampace: dynamic force-coordination trainer for the upper extremities" by Arno et al. in IEEE 10th International Conference on Rehabilitation Robotics 2007; **(c)**: Parallel Planar type CDRR - Planar Cable-Driven Robot [52] ©2014 IEEE. E/Reprinted, with permission from "Workspace analysis of upper limb for a planar cable-driven parallel robots toward upper limb rehabilitation" by XueJun et al. in IEEE International Conference on Control, Automation and Systems 2014; **(d)**: Serial Planar type CDRR - MEDARM [11] ©2007 IEEE. E/Reprinted, with permission from "A planar 3DOF robotic exoskeleton for rehabilitation and assessment" by Stephen et al. in 29th Annual International Conference of the IEEE Engineering in Medicine and Biology 2007; **(e)**: Parallel End-effector type CDRR - PACER [54] ©2015 IEEE. E/Reprinted, with permission from "Modeling and control of a novel home-based cable-driven parallel platform robot: PACER" by Aliakbar et al. in IEEE/RSJ International Conference on Intelligent Robots and Systems 2015; **(f)**: Serial End-effector type CDRR-Active therapeutic device (ATD) [55] ©2014 IEEE. E/Reprinted, with permission from " A Damper Driven Robotic End-Point Manipulator for Functional Rehabilitation Exercises After Stroke" by Ard et al. in IEEE Transactions on Biomedical Engineering 2014.

**Human-robot physical interaction:** Exoskeleton type CDRRs for upper limb rehabilitation have mechanical structure correspondence to the upper limb skeleton structure of a human extremity. The CAREX [8] and Dampace [50] are parallel and serial exoskeleton type CDRRs, respectively, shown in Figs. 2-(a), (b). The CAREX [8] is a 5-DOF rehabilitation robot. It has a novel design, in which cuffs of CAREX are attached to the upper arm that are driven by the cables. The Dampace [50] is a passive exoskeleton to assist the 3 rotational DOFs of shoulder joint and one DOF of elbow joint. It was designed to combine the functional training of daily life activities with force coordination training.

A parallel and one serial planar type CDRRs are shown in Figs. 2-(c) and 2-(d), respectively. They provide move-ments in a specific plane but allow limb to move in three-dimensional space. Planar Cable-Driven Robot [51], [52] is a parallel 3-DOF CDRR for upper limb rehabilitation. It provides a relatively large workspace and less moving inertia. MEDARM [11] is a serial CDRR and can assist in the horizontal plane. The novelty of this design is a curved track that allows the independent control of a patient's joints.

A parallel and serial end-effector type CDRRs are shown in Figs. 2-(e) and 2-(f), respectively. PACER [53], [54] is a parallel end-effector type CDRR and consists of a cable driven antagonistically actuated prismatic joint. This novel design was presented to facilitate the home-based rehabilitation device. Active therapeutic device [55] is a serial end-effector type CDRR with 3-DOF. It was designed to support the functional reaching movements.

**Human-Machine Interface (HMI)**: The IntelliArm [86], Dampace [50], L-EXOS [87], Planar CDPR [51] [52], HIT-Glove [88], MULOS [89], and Sophia-3/4 [90] are few upper limb CDRRs that benefited from a type of HMI to train the patients. The IntelliArm rehabilitation robot reported a simple virtual game interface to keep patient motivated and enhance the training ability, in which the movement of the cursor in the game was commended from the patient's joint movement. The Dampace rehabilitation robot employed a racing game interface to increase the recovery rate, where the good driving control in the game requires good coordination of elbow and shoulder torques. The L-EXOS robot incorporated the cubes game to train the patient, in which patient was commanded to move the cubes in the virtual scenario. The Planar CDRR [51] used three virtual rehabilitation therapy games: a) painting game, b) pong game, and c) ball game to train the patient in virtual environment. The HIT-Glove used tactile interface, where a display screen interface with force sensors was used to measure the patient's effort in the training exercise, while MULOS CDRR deployed a specially designed 5-DOF joystick.

**Clinical Study of CDRRs for upper limb:** Although, a large number of CDRRs for upper limb rehabilitation are reported in literature, only a few of them have gone under a clinical investigation. RUPERT [81], [82], [83] and Wearable Rehabilitation Robotic Hand [84] have reported a successful clinical study at Level 1. The RUPERT carried out the clinical study with eight able-bodied volunteers. The focus of the study was to check the robot performance in actual rehabilitation environment. The Wearable Rehabilitation Robotic Hand CDRR performed the clinical study with only one healthy subject. L-EXOS [87], Tendon Driven Hand Orthosis [66], and CAREX [8] are three CDRRs that have their clinical study reported at Level 2. The L-EXOS went through the clinical trial at the Neurorehabilitation Unit, at the University of Pisa. It was also integrated with the virtual environment to motivate the patients in the rehabilitation process. The preliminary clinical trial was carried out with a 60-year-old patient and reported satisfactory results. They also examined the robot performance by conducting an extended clinical trial with six patients for six weeks. The CAREX demonstrated the clinical study with eight healthy volunteers and one stroke patient, and found the subject efficiently followed the desired trajectory. The CAREX does not cause hinders or incorrect posture to patient movements, except that they felt a little tired after one hour of exercise. NeReBot [91] is one of the CDRRs with results of clinical study at Level 3. This robot was used in two clinical tests. In the first scenario, the NeReBot was used in addition to the traditional method of treatment/therapy [92]. Thirty-five patients participated in this clinical trial and significant recovery was reported. In the second scenario, the NeReBot was partial substitution to the standard rehabilitation treatment by using a dose-matched approach [93]. Thirty-four patients, 11 females and 23 males, participated in this clinical trial, and patients show significant degree of acceptance to the robotic training. Up to now, none of CDRRs has reached level four of clinical trial.

### B. CABLE-DRIVEN ROBOTS FOR LOWER LIMBS

The neuro-rehabilitation of the lower limb, especially for the gait training, demands a great amount of time and effort from the physiotherapists. In most the cases, more than two physiotherapists are required to assist the rehabilitation training of patient with lower limb disabilities. CDRRs for lower limb rehabilitation can reduce the burden on the health care, cost, and can increase the recovery rate [94]. Contrary to CDRRs for the upper limb, less attention has been paid in designing CDRRs for the lower limb.

Few CDRRs for lower limb rehabilitation are LOPES [95], [96], [97], C-ALEX [98], COWALK-Mobile 2 [99], Lower Limb Rehabilitation Robot [100], and Bio-inspired Soft Wearable Robot [101]. A comprehensive list of existing CDRRs for lower limb are summarized in Table 8.

**Target movements for rehabilitation:** Most of the lower limb CDRRs are designed to assist all the joints of the lower limb; hip, knee and ankle joints. There are however some CDRRs to assist only one or two lower limb joints, such as, Powered Ankle Prostheses [102] and C-ALEX [98] that are proposed to assist the rehabilitation of ankle joint and hip-knee joints, respectively. The categorization of lower limb CDRRs based on number of target joints for rehabilitation is shown in Table 8.

**Types of actuators:** The CDRRs for the lower limb can be classified into three types: electric, pneumatic, and series elastic actuator based on how the energy is provided to the actuators. As can be seen from the Table 8, most of the CDRRs for the lower limb used the electric and pneumatic actuators. LOPES [95], [96], [97] is the only CDRR for the lower limb, which used series elastic actuator.

**Human-robot physical interaction:** Referring to how CDRRs can be connected to the lower limb, they can be categorized into four categories, namely, serial exoskeleton, parallel exoskeleton, planar and parallel end-effector types as shown in Fig. 3. LOPES [95]–[97] is a serial exoskeleton type CDRR, which combines a pelvis segment and a leg. This design allows two therapy protocols, patient in charge and robot in charge. C-ALEX [98] is a parallel exoskeleton

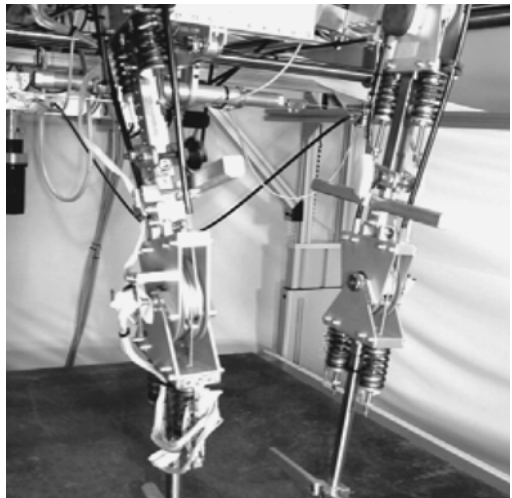

(a) Serial Exoskeleton CDRR - LOPES [96] ©2007 IEEE

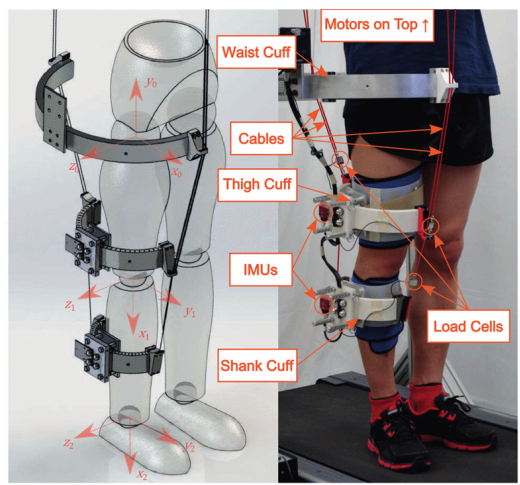

(b) Parallel Exoskeleton CDRR - C-ALEX [98] ©2015 IEEE

FIGURE 3: Examples of cable driven rehabilitation robots for Lower limb. **(a)**: Serial Exoskeleton type - LOPES [96] ©2007 IEEE. E/Reprinted, with permission from "Design and Evaluation of the LOPES Exoskeleton Robot for Interactive Gait Rehabilitation" by Jan et al. in IEEE Transactions on Neural Systems and Rehabilitation Engineering 2007; **(b)**: Parallel Exoskeleton type - C-ALEX [98] ©2015 IEEE. E/Reprinted, with permission from "Design of a cable-driven active leg exoskeleton (C-ALEX) and gait training experiments with human subjects" by Xin Jin et al. in IEEE International Conference on Robotics and Automation 2015.

type CDRR designed for the gait training. It has a simple structure and add less moving inertia to the human limbs. Another CDRR presented in [103], [104] is a planar cable-driven parallel robot driven by four cables. It was proposed for gait rehabilitation and can produce a wrench-closure trajectory. Table 8 summarizes the CDRRs for the lower limb rehabilitation and shows the type of interaction they have with patient's joints. Advantages and disadvantages of these types of physical interactions are discussed at section III-D.

**Human-Machine Interface:** Contrary to CDRRs for the upper limb, limited information is reported in literature on human-machine interfaces for the lower limb training. The C-ALEX [98] was supported by a display screen to show the subject's leg movement in sagittal plan. The LOKOIRAN [105] and String Man [106] have also benefited from human-machine interface to motivate patients for exercises.

**Clinical study:** Similar to CDRRs for the upper limb rehabilitation, the clinical study of CDRRs for the lower limb rehabilitation has received very limited attention. Only a few CDRRs, LOPES [107], Soft Exosuit [108], String Man [106], Locomotor [109], and TPAD [110] were able to support their findings with clinical studies. The LOPES [107] carried out the clinical study with ten patients, and significant improvement in the rehabilitation process was reported. The Soft Exosuit [108] performed the clinical study with three patients, and showed a satisfactory performance in clinical trial. The String Man [106] has performed a clinical study with both dummies and healthy people. The Locomotor [109] used the 11 subjects with incomplete spinal cord injury to conduct the clinical trial. The TPAD [110] recruited the seven healthy subjects to perform the clinical trial. Based on the classification mentioned in the table 3, LOPES, Soft Exosuit, and Locomotor have their clinical studies at level 2, whereas the String Man and TPAD have clinical study at level 1.

### C. CABLE-DRIVEN ROBOTS FOR WAIST

The last group of CDRRs targets patients with waist injuries, Contrary to CDRRs for upper limb or lower limb rehabilitation, limited work has been done in developing CDRRs for the waist rehabilitation. The existing CDRRs are HWRR-Waist Rehabilitation Robot [111], [112], CDPR [113], CPRWR [114], and CDPRR [115].

Table 7 shows list of CDRRs for waist rehabilitation and their features with more detail. All of the existing CDRRs for waist rehabilitation are parallel exoskeleton type. The electric and pneumatic types of actuators are used to drive the CDRRs for waist rehabilitation. None of the CDRR for waist rehabilitation went under the clinical investigation or provided the human-machine interface.

### D. DISCUSSIONS ON DESIGN CHARACTERISTICS

The cable-driven rehabilitation robots are aimed to assist patient improve their ability in performing the daily life activities. As such, the CDRRs are typically designed to support multiple target movements of upper or lower limbs. However, this is challenging for CDRRs given the uni-lateral actuation nature of the cables. The complexity in the kinematic, static, and dynamic analysis of CDRRs increases with the increases in the number of target rehabilitation movements. Additionally, actuators in CDRRs are usually placed at the base, and joints are actuated through cables, so it is challenging to design a controller that can assist large numbers of joints.

Concerning the energy source of an actuator, electric actuators have excellent motion control capabilities, accuracy

TABLE 4: CDRRs for upper limb rehabilitation: targeting only a single joint

| Name of CDRR; Reference | DOF | Actuator | Sensor | HRPI | Control Strategy | Clinical & HMI |
|---|---|---|---|---|---|---|
| *Target movements of rehabilitation; shoulder movements* | | | | | | |
| MEDARM [11] | {3} | EA | Encoder | PLS-E | Joint position control | Prototype |
| Upper Arm Exoskeleton [116] | {3} | —— | —— | PE | Force control | —— |
| Wearable Soft Orthotic Device [58] | {1} | SEA | Piezoresistive flex + IMUs | PE | Impedance control | Prototype |
| Active Soft Orthotic Device [60] | {1} | SEA | Electromagnetic + 6-axis force-torque | PE | Adaptive control | Prototype |
| *Target movements of rehabilitation; elbow movements* | | | | | | |
| NEUROExos [61] | ⟨4⟩ | EMA+HA | Encoder + Load cells + Linear potentiometers | SE | Passive-compliance + Impedance torque ctrl | Prototype |
| BJS [63] | {1}+⟨3⟩ | SEA | Rotary + Hall effect sensors | SE | Sliding mode controller | Prototype |
| Upper Limb Soft Exoskeleton [117] | {1} | EA | Flex + Film pressure sensor | PE | Hierarchical cascade controller | Prototype |
| Adaptive Rehab. Sys. [118] | {4} | PAM | KinectTM + EMG sensors | SE | Pressure control | Prototype |
| Elbow Exoskeleton [85] | {2} | SEA | Torque sensors | PE | PID feedback control | Prototype; Forearm |
| CADEL [119] | {1} | EA | Force + IMU sensors | PE | —— | Prototype |
| Elbow Rehab. [120] | {1} | EA | —— | PE | Sensor-assisted control | Prototype |
| *Target movements of rehabilitation; forearm movements* | | | | | | |
| ATD [55] | {3} | EA | Six-DOFs force sensor + Angular encoders | SEE | Impedance + Position control | Prototype |
| *Target movements of rehabilitation; wrist movements* | | | | | | |
| CDWRR [64] | {3} | EA | Load + IMU | PE | Assist-as-needed | Prototype |
| *Target movements of rehabilitation; fingers movements* | | | | | | |
| Wearable Rehab. Robo. Hand [84] | {2} | PAM | Angle + Pressure + Force Sensor | PE | Variable integral PID control | Prototype; CS-L1 |
| CADEX Glove [121], Finger Rehab. [122] | Multi-DOF | EA | —— | PE | —— | Prototype |
| Wearable Hand [123] | —— | EA | Load cell | PE | Force control | Prototype |
| Tendon Driven Hand orthosis [66] | {3} | EA | Encoder + Load sensor | PE | Position PID controller | Prototype; CS-L2 |
| SNU Exo-Glove [124] | Multi-DOF | EA | Custom-made tension + Force sensors | PE | Isotonic + Isokinetic + Impedance control | Prototype |
| CAFÉ [125] | {3} | EA | Force + Angle sensors | SE | PI control | Prototype |
| Hand Exoskeleton [126] | {7} | EA | —— | SE | —— | Prototype |
| Hand Exoskeleton [127] | {4} | EA | Force sensors | SE | Resistance compensation control | Prototype |
| HIT-Glove [88] | {2} | EA | Position + Force sensors | SE | Patient-cooperative control | Prototype; HMI:DS |
| Hand Exoskeleton [128] | Multi-DOF | EA | —— | SE | High-order iterative learning Ctrl (ILC) | Prototype |
| IOTA [70], [129] | {2} | EA | Encoders + Bend sensors | SE | Proportional control | —— |
| Hand Exoskeleton [130] | Multi-DOF | EA | Hall + EMG sensors + Bend sensors | SE | Active disturbance rejection control | Prototype |
| Hand Exoskeleton [131], Soft Glove [132] | Multi-DOF | EA | EMG sensors | SE | —— | Prototype |

First column: shows the CDRR name and reference of corresponding publication. Second column: indicates the active {*No*} and passive ⟨*No*⟩ DOFs of CDRR; Third column: shows the corresponding sensors Fourth column: indicates the type of CDRR; PE- Parallel exoskeleton, PLS-E: Planar Serial Exoskeleton, SEE: Serial End-effector, and SE: Serial Exoskeleton. Fifth column; describes the type of control strategy. Sixth column; refers of type of actuator; EA - Electric actuator, SEA -series elastic actuator, EMA: Electro Magnetic Actuator, and PAM - Pneumatic artificial muscle. Last column; provides the more information about: prototype availability, clinical studies level, and Human-Machine Interface; CS-L1: Clinical Study at Level 1, CS-L2: Clinical Study at Level 2, and HMI:DS: Human-Machine Interface: Display Screen.

TABLE 5: CDRRs for upper limb rehabilitation: targeting any combination of: shoulder, elbow, forearm, wrist and fingers

| Name of CDRR; Reference | DOF | Actuator | Sensor | HRPI | Control Strategy | Clinical & HMI |
|---|---|---|---|---|---|---|
| **Target movements of rehabilitation; shoulder + elbow movements** | | | | | | |
| Dampace [50] | ⟨5⟩ | EA + HA | ——- | Adjustable SE | Feedback control | Prototype; HMI:RG |
| Upper Limb Exoskeleton [133] | {5} | EA | ——- | SE | Intention-driven robotic control | ——- |
| MEDARM [11] | {3} | EA | Encoder | PE | Joint Position Ctrl | Prototype |
| Exoskeleton Rehab. Robot [78] | {5}+⟨5⟩ | PAM | Optical motion capture system | SE | EMG based real-time control | ——- |
| ABLE [134] | {4} | EA | 3D displacement + Position sensors | SE | PID-position control | Prototype |
| 9-DoFs Rehab. Robot [79], [80] | {8}+⟨1⟩ | PAM | Angle + Force sensors | SE | PID Position + Force control | Prototype |
| CARR-4 [135] | {4} | EA | Load + IMUs sensors | PE | Robust Adaptive ILC + Feed forward Ctrl | Prototype |
| Auxilio [136] | {3} | DC-EA | Kinect motion sensor | PE | Slack mitigation Ctrl | Prototype |
| Parallel Structure 4-2 [137] | Multi-DOF | DC-EA | Encoder | Sus-PEE | Speed + Position closed-loop control | Gait rehab |
| Upper Limb Rehab. Robot [138] | {4} | DC-EA | IMU + Encoder | SE | PID control | Prototype |
| L-EXOS [87] | {4}+⟨1⟩ | DC-EA | Position sensor | SE | Impedance control | HMI:VE; CS-L2 |
| CDE [139] | {4} | EA | Load sensors | PE | Stiffness oriented control | Prototype |
| **Target movements of rehabilitation; Wrist + Fingers movements** | | | | | | |
| MAHI [140] | {3} | EA | Angle sensors | SE | PD control | Prototype |
| **Target movements of rehabilitation; elbow + wrist movements** | | | | | | |
| ULERD [71], [75] | {3}+⟨4⟩ | EA | Inertia sensor | SE | PID control | HMI:VE |
| **Target movements of rehabilitation; shoulder + elbow + wrist movements** | | | | | | |
| IKO [76], [77] | {5}+⟨3⟩ | EA+PAM | Pressure + Length measuring sensors | SE | Position + Rotation control | Prototype |
| X-Arm-2 [141] | {8}+⟨6⟩ | EA | Torque sensor | PE | Joint-impedance | Prototype |
| CABexo [73] | {6} | EA | ——- | SE | ——- | ——- |
| CABXLexo-7 [142] | {7} | EA | ——- | SE | ——- | Prototype |
| RUPERT [81]–[83] | {4} | PAM | Position + Inertial + Pressure sensors | SE | Open-loop feed forward position control | Prototype; CS-L1 |
| Planar Cable-Driven Robot [51], [52] | Multi-DOF | EA | 6-DOF force sensor | PLP-EE | Position-Based Admittance Control | Prototype; HMI:VG |
| CADEN 7 [143] | {4} | EA | Redundant position sensors | SE | Position + Force impedance Ctrl | Prototype |
| Robotic Exoskeleton [144] | {4} | PAM | Peripheral sensor | SE | Adaptive fuzzy sliding mode control | Prototype |
| UP Exoskeleton [145] | {4} | EA | Load + IMU sensor | PE | feedforward control | Prototype |
| **Target movements of rehabilitation; shoulder + elbow + forearm movements** | | | | | | |
| MULOS [89] | {5} | EA | ——- | SE | Vel. + Force PID | HMI:JCS |
| CAREX [8] | {5} | EA | Orientation + Rotary + Load sensors | PE | Assist-as-needed | Prototype; CS-L2 |

keywords: { }: active DOF and ⟨ ⟩: passive DOF, PE: Parallel Exoskeleton, SE: Serial Exoskeleton, Sus-PEE: Suspended Parallel End-effector, PLP-EE: Planar Parallel End-effector, DC: Direct current, EA: Electric Actuator, HA: Hydraulic Actuator, PAM: Pneumatic Artificial Muscle, VE: Virtual Planar Environment, VG: Virtual Games, JCS Joystick Control System, CS-L1: Clinical Study at Level 1, and CS-L2: Clinical Study at Level 2.

TABLE 6: CDRRs for upper limb rehabilitation: targeting all joints of arm except fingers

| Name of CDRR; Reference | DOF | Actuator | Sensor | HRPI | Control Strategy | Clinical & HMI |
|---|---|---|---|---|---|---|
| **Target movements of rehabilitation; shoulder + elbow + forearm + wrist movements** | | | | | | |
| SUEFUL-7 [72] | {7} | EA | Force + Torque sensors | Adjustable SE | Impedance control | Prototype |
| IntelliArm [86] | {7}+⟨2⟩ | EA | Force + Torque sensors | Adjustable SE | Intelligent stretching + Zero resistance regulation control | Prototype; HMI: G |
| Soft-Actuated Exoskeleton [146] | {7} | PAM | High linearity + Torque + Pressure sensors | SE | Impedance control + PID control | Prototype |
| CAREX-7 [9], [74] | {7} | EA | Load + IMU sensors | PE | Assist-as-needed | Prototype |
| Upper-Limb Powered [147] | {7} | Brushed EA | Potentiometer + Encoder | SE | Position + Force-impedance control | Prototype |
| Rehab. Robot [148] | Multi-DOF | EA | Force sensors | PLEE | ——- | Prototype |
| Sophia-3 and Sophia-4 [90] | Multi-DOF | DC-EA | Optical encoders | PLP-EE | Impedance + Model-based adaptive Ctrl | Prototype; HMI:VE |
| Cable-driven Rehab. Robot [149] [150] | Multi-DOF | ——- | Encoder | Sus-EE | Sliding Mode Tracking Control | Prototype |
| NeReBot [91], [151], [152] | {3} | EA | ——- | Sus-EE | PCIO4 CONTROL | Prototype; CS-L3 |
| MariBot [153] | {5} | EA | Incremental encoder | Sus-EE | PID position control | Prototype |
| MACARM [154] | {6} | EA | 6-DOF load sensor | Sus-EE | Motion control | Prototype |
| PACER [53], [54] | {6} | Linear EA | ——- | Spatial+ Planar-EE | Feedback + Admittance control | ——- |
| CDULRR [155] | {3} | ——- | Force | PEE | Assist-as-needed | Prototype |
| Mirror-Image Motion Dev. [156] | ——- | BLDC-EA | Inertial Sensors | ——- | PD control | Prototype |
| CDRR Exoskeleton Sys [156] | ——- | DC-EA | IMU Sensors | SE | PID force feedback control | Prototype |

Keywords: { }: active DOF, ⟨ ⟩: passive DOF, SE: Serial Exoskeleton, PE: Parallel Exoskeleton, PLP-EE: Planar Parallel End-effector, Sus-EE: Suspended End-effector, DC: Direct current, EA: Electric Actuator, and PAM: Pneumatic Artificial Muscle, HMI:G: Human-Machine Interface: Cursor Positing Game, HMI:VE: HMI:Virtual Planar Environment, and CS-L3: Clinical Study at Level 3.

TABLE 7: CDRR for waist rehabilitation

| Name of CDRR; Reference | DOF | Actuator | Sensor | HRPI | Control Strategy | Clinical & HMI |
|---|---|---|---|---|---|---|
| HWRR-Waist Rehab. Robot [111] | {2} | PAM | Position + Orientation + Tension sensors | PE | Coordinate control | ——- |
| CDPR [113] | ——- | PAM+EA | Wire displacement + Tensile force sensors | PE | PID + fuzzy control | Prototype |
| CPRWR [114] | {3} | PAM | Wire displacement + Tension + Laser + Vision sensors | PE | ——- | ——- |
| CDPRR [115] | {3} | EA | Force sensors | PE | ——- | LO Rehab |

All the CDRR for the waist rehabilitation are presented in this table. Keywords: PE- Parallel exoskeleton, EA - Electric Actuator, PAM - Pneumatic Artificial Muscle, and LO Rehab - Also, provide the lower limbs rehabilitation.

TABLE 8: CDRRs for Lower limb rehabilitation

| Name of CDRR; Reference | DOF | Actuator | Sensor | HRPI | Control Strategy | Clinical & HMI |
|---|---|---|---|---|---|---|
| **Target movements of rehabilitation; Hip movements** | | | | | | |
| Hip-only Soft Exosuit [157] | —–- | EA | Load cells + IMUs | PE | Force-based position control | Prototype |
| Multi-Robotic Sys. [158] | {2} | EA | Tension + Position sensor | PEE | Closed-loop speed + PID control | {6} Pelvic module |
| **Target movements of rehabilitation; knee movements** | | | | | | |
| Soft Wearable Robot. [159] | {1} | Flat PAM | —–- | PE | —–- | Prototype |
| 4-4 CDPR [103], [104] | {3} | EA | —–- | PLP-EE | PD control | Gait rehab Shank Att. |
| **Target movements of rehabilitation; ankle movements** | | | | | | |
| Soft Parallel Robot [160] | {3} | PAMs | —–- | PE | —–- | Prototype |
| Soft Exosuit: ankle [108] | —–- | EA | Load cells + IMUs | PE | hierarchical control | CS: L2 |
| Powered Ankle Prostheses [102] | {3} | EA | IMUs | PE | Impedance control | Prototype |
| AFO [161] | {1} | PAM | Encoder | SE | Iterative learning control | Prototype |
| Bio-inspired device [101] | Multi-DOF | PAM | Strain + IMUs + Pressure sensors | PE | LTI controller | Foot Rehab |
| CABLEankle [162] | {3} | EA | —–- | PE | —–- | Prototype |
| Ankle exoskeleton [163] | Multi-DOF | BLDC-EA | Force + IMU sensors | PE | Feedforward + PD control | Prototype |
| Polycentric ankle exoskeleton [164] | ⟨2⟩ | EA | Encoder + IMU sensors | SE | Feedforward + PID Neural Network control | Prototype |
| **Target movements of rehabilitation; hip + knee movements** | | | | | | |
| LOPES [95] [96] [97] | {8} | SEA | 6D force + Position sensor | SE | Impedance control | CS:L2 |
| Soft exosuit [165] | —–- | EA | load cells + IMUs | PE | Adaptive control | Prototype |
| C-ALEX [98] | {2} | Servo EA | IMU + Load sensors | PE | Assist-as-needed control | HMI:VE |
| **Target movements of rehabilitation; hip + knee + ankle movements** | | | | | | |
| Powered Lower Limb Orthosis [166] | {8}+⟨2⟩ | High PAM | Position + Pressure + Torque sensors | SE | 3-level-PID joint torque control | Prototype |
| XoR [167] | {8}+⟨4⟩ | PAM | Force + EMG sensors | SE | Impedance control | Prototype |
| Hip, Knee and Ankle Module [168] | {1}+{2}+{1} | PAM | —–- | SE | Proportional myoelectric control | Prototype |
| Soft-exosuit [169] | {6} | PAMs | —–- | PE | Time based control | Prototype |
| Lower limb RR [100] | —–- | DC-EA | Encoder + Load + Stretcher | PEE | Impedance control | Prototype |
| CDRR [170] | —–- | DC-EA | —–- | PEE | Torque Control | Shank Att. |
| Underactuated Lower Limb Exoskeleton [171] | {3}+⟨10⟩ | DC-EA | Force + Position sensors | SE | Admittance control | Assist 1-DOF spine |
| ROPES [172], [173], [174] | Multi-DOF | DC-EA | Load + Orthosis + IMU sensor | SE | Joint-space PD + Task-space force control | Gait rehab |
| WDS [175] [176] | {4} | EA | Load cell+switch + Encoder | PEE | Open-loop control | Shank Att. |
| LOKOIRAN [105] | ⟨9⟩ | EA | Force sensors | (P+S)E | Admittance control | HMI:VRE |
| String Man [106] | {6} | EA | Tension + Force + Position sensors | PE | Impedance control | CS:L1 HMI:VE |
| TPAD [110] | —–- | EA | Tension sensors | PE | Assist-as-needed control | CS:L1 |
| CD Locomotor Sys. [109] | —–- | EA | 3D position sensors | PE | Resistance control | CS:L2 |
| CUBE [177], [178] | {5} | EA | —–- | PE | Position feedback control | UP Rehab |
| WeARS [179] | —–- | EA | Load cell | SE | Resistance control | Gait rehab |
| Rigid-Soft Hybrid Exoskeleton [180] | multi-DOF | EA | Torque sensor | PE | Position/force closed-loop control | Gait rehab |

Keywords: { } - active DOF, ⟨ ⟩ - passive DOF, PE- Parallel Exoskeleton, PEE - Parallel End-Effector, PLP-EE - Planar Parallel End-Effector, SE - Serial Exoskeleton, AC - alternating current, DC - Direct Current, EA - Electric Actuator, SEA - Series Elastic Actuator, PAM - Pneumatic Artificial Muscle, CS-L1 - clinical study at level 1, Shank attachment - CDRR is attached to the patient shank, Foot Rehab - CDRR is also providing the foot rehabilitation, Gait rehab - CDRR is providing the gait rehabilitation, Assist 1-DOF spine - CDRR is also providing the 1-DOF spine rehabilitation , and HMI:VRE - HMI; virtual reality environments.

and repeatability for the CDRRs. Additionally, they are quieter than hydraulic and pneumatic actuators. However, some electric actuators have backlash and tend to overheat when holding a CDRR's joint in a locked position. The pneumatic actuators are shock, explosion, and spark proof. However, their use in CDRRs is problematic particularly when those are operating at low pressure in rehabilitation applications that requires low force and slow speed. CDRRs with hydraulic actuators can deliver large forces and has ability to handle shock loads. However, the drawbacks of hydraulic actuators are fluid leakage and requirement for regular maintenance. Compare to electric and pneumatic actuators, the use of hydraulic actuators in CDRRs has received very limited attention has so far only used in two CDRR designs.

In relation to the types of human-robot physical interaction, serial type exoskeleton CDRRs have several advantages, and one can apply traditional rigid link serial robot modelling and control strategies for this type of CDRRs but they need tuning and readjustment for each patient. The misalignment between the corresponding revolute axes of a serial exoskeleton CDRR and anatomical axes of human extremities must be avoided. Unlike serial CDRRs, parallel ones don't have rigid linkages and therefore there are no misalignment issues. However, traditional cable-driven parallel mechanism modelling and control strategies are not directly applicable as those methods do not model the collision between cables and segments of patients. The limited workspace is another significant challenge for designing a parallel type exoskeleton CDRR. Serial end-effector type CDRRs, have drawbacks of controllability and small workspace, which limit their use in rehabilitation applications. Parallel end-effector CDRRs, don't have the adaptability and misalignment issue between the robot and patients, as they don't have the revolute joint or rigid-link. But similar to parallel exoskeletons, they also have limited workspace.

A significant number of CDRRs use the human-machine interface to provide rehabilitation training. HMIs mainly provides the following major advantages:

- ☐ It increases the patient interest and attention in the rehabilitation exercise.
- ☐ It keeps the patient motivated during rehabilitation.
- ☐ Allows selecting the training scenarios based on the patient's interest and monitoring of the training data.
- ☐ It is not only leverage patient's interaction with CDRR but also increase the rate of recovery as well.

Although the design and development of CDRRs has received significant attention in the last two decades, less work has been undertaken to verify the performance of these robots in actual rehabilitation settings. Among these work, most of the CDRRs are at level 1 or 2 of clinical study and few in level 3. The existing trials however reported the effectiveness and patient's acceptance of CDRRs in rehabilitation training. However, there is still very limited amount of clinical study, to move CDRRs from technical laboratories to rehabilitation centers or hospitals.

## IV. CONTROL STRATEGIES FOR CDRRS

The main control strategies developed for cable-driven rehabilitation robots are discussed in this section. The terminology is mainly based on the one proposed in [43], which is summarized in Table 9. We review the control strategies and techniques in CDRRs into five main categories including impedance-based, PID-based, admittance-based, assist-as-needed, and adaptive based controllers, and a sixth group comprising the rest of controllers. Tables 10 and 11 summarize important information for different CDRRs focusing on the controller's type, level, modelling approach, type of performance analysis, operating space, and stability. Moreover, the overall schematics of those controllers are depicted in Fig. 4. Finally, the advantages and disadvantages of control strategies are summarised at the end of this section.

The level of strategy to control individual components or the overall CDRR can be classified into low, mid, and high level, concerning joint or task spaces, inverse kinematic or dynamic, and path planning, respectively. Different control techniques can be; model-based that rely on analytical models, model-free that use machine learning techniques, empirical methods, or hybrid which is a combination of model-based and model free approaches. The performance of the controllers is usually evaluated in literature based on the joints angle tracking error, Cartesian error for task space tracking, or position tracking error. Task space or configuration space are the main two operating space used in controlling CDRRs. Task space refers to the position and orientation of the end-effector in Cartesian space. Configuration space refers to the set of all possible configurations of CDRRs' joint. The stability of a control strategy means it's ability to produce bounded outputs with bounded inputs.

### A. IMPEDANCE CONTROL STRATEGY

Impedance control is a kind of assistive control strategy, which makes the rehabilitation tasks easier and safer to accomplish, while supporting more repetitions. There are two main categories of impedance control strategies utilized in the CDRRs: Force-based and EMG-based impedance control. Fig. 4 (a) presents the overall block diagram of an impedance control system, in which load cells or EMG sensors are alternatively deployed in the closed loop of a force-based or EMG-based controller. Impedance parameters could be also adjusted in real time as a function of upper and lower limb posture, as discussed in [72]. There is a large number of research on the use of impedance based control strategy in CDRRs, as summarized in Table 10.

#### 1) Force-based impedance control strategy
Force-based impedance control is a dynamic control that relates the position of the patient body to the corresponding force feedback provided by the assistive robot. In this strategy, the patient is supposed to follow a particular reference trajectory. Once the patient deviates from the desired trajectory, a restoring force is applied on the patient by the assistive robot. The amount of restoring force is directly proportional to the deviation between actual and reference trajectories. Normally, deviation from a desired trajectory is allowed up to some margin before restoring force is applied. As shown in Fig. 4 (a), the error between actual and reference trajectories is first provided to a virtual impedance block before the force controller. For safety reasons, the desired interaction force is adjusted in virtual impedance, via therapist inputs, to ensure actual interaction forces between CDRR and patient remains below certain thresholds. The output of virtual impedance is then fed to the force controller block, from where the torque/force are commended to the robot.

TABLE 9: Glossary of terms: control strategies

| Term | Description |
| --- | --- |
| Impedance based controller | is based on force feedback as a function of the position measured. |
| Position/Orientation PID controller | is based on a control loop feedback to regulate position, velocity, force and other variables. |
| Admittance based control | modulates the position trajectory as a function of the force measured. |
| Assist-as-needed based controller | modulates the position trajectory as a function of the force measured. |
| Performance based adaptive controller | is based on measuring patient performance and preceding repetitions. It also adapts some aspects (e.g. force, stiffness, time, path) of assistance. |

**Upper limb:** The effectiveness of impedance based control strategy in CDRRs is demonstrated by several works, including NEUROExos [61], Active therapeutic device (ATD) [55], X-Arm-2 [141], Sophia-3 and Sophia-4 [90], and SUEFUL-7 [72] via experimental investigation. Two different strategies are proposed in [61], one for robot-in-charge mode, which is an independent control of joint positions and one for the patient in-charge mode, that is near-zero impedance torque control. L-EXOS [87], Sophia-3, and Sophia-4 [90] deployed impedance based control strategies with the assistance of HMI. However clinical investigation of the impedance based controllers for upper limb rehabilitation is hardly available in literature. L-EXOS [87] is the only CDRR with clinical result of an impedance based controller, which is based on trials performed with a single patient.

**Lower limb:** the use of impedance based control strategies in CDRRs assisting patients with the lower limb disabilities is also reported by [100], Powered Ankle Prostheses [102], String Man [106], and LOPES [95], [96], [97].

2) **EMG-based impedance control strategy**

It is a kind of muscle status dependent control strategy. It encourages patient's effort by using the patient's EMG signals to proportionally control the assistance. EMG signals have the information regarding muscle activation, which is important for rehabilitation process and controlling CDRRs. The EMG signal are normally used in two ways in a control system: for activation of the robot, or as a feedback for adjusting the controller. [181]. Fig. 4 (a) shows the block diagram of a EMG-based impedance control strategy. It consists of four stages [182]: first, signal pre-processing, which involve data acquisition and signal enhancement, second feature extraction, third dimensional reduction by removing the non-distinguishing feature, and fourth, classification of muscle activation patterns. Then, the estimated pattern of muscle contraction is fed to the controller block.

**Upper limb:** The application of EMG-based control systems is also reported in few upper limb CDRRs such as Exoskeleton Rehabilitation Robot [78] and XoR [167].

### B. PROPORTIONAL INTEGRAL DERIVATIVE CONTROLLER (PID)

PID controllers are found in several CDRRs, as they are robust to wide range of rehabilitation process conditions and consists of a very simple structure. PID control systems are well-known for regulating position, velocity, force and other process variables, via using control feedback loop. The controller response is characterized by settling time, overshoot, rise time, and steady state error.

Fig. 4 (b) depicts a block diagram of a simple PID control strategy. This control system consists of three fundamental parameters [183]. The proportional gain determines the ratio of system response to the error between a set point and process variable. The integral gain is tuned to minimise steady state error. The derivative term tries to bring the rate of change of error to zero and reduces the overshoot. The CDRRs used PID control are summarized in Table 10.

**Upper limb:** several experimental investigation and performance analysis of PID position and velocity controllers for CDRRs are available in literature, such as ULERD [71], [75], MULOS [89], ABLE [134], 9-DoFs rehabilitation robot [79], [80], NeReBot and MariBot [91], [151], and Tendon Driven Hand orthosis [66]. PI and PD controllers are also used in CDRRs for controlling the position of patients hands, CAF [125] and MAHI Open Wrist exoskeleton [140]. Clinical verification of PID position controllers is reported for NeReBot [91], [151], and Tendon Driven Hand Orthosis [66]. Clinical trials reported significant improvement and patients have shown favourable impressions about the treatment, without any side effects. However, the sole use of position control is not enough to ensure safe dynamic interaction between human and robot [184] as it does not take into account the required force.

**Lower limb and Waist:** the PID control strategy is applied in few lower limb rehabilitation works such as Multi-Robotic System [158] and Powered Lower Limb Orthosis [166]. It is also used in waist rehabilitation by [113], in which a fuzzy PID control is developed and examined.

### C. ADMITTANCE BASED CONTROL

It relies on the measurements of interaction force to ensure the compliance of robots during rehabilitation. Admittance control strategy is the opposite of impedance control method. While admittance-based strategy controls motion by measuring the force, the impedance-based strategy controls force by measuring the motion and deviation from a referenced trajectory [185]. Fig. 4 (c) depicts the overall block diagram of admittance based control system. It is assumed that patient is subject or subjected to interaction or external forces. The

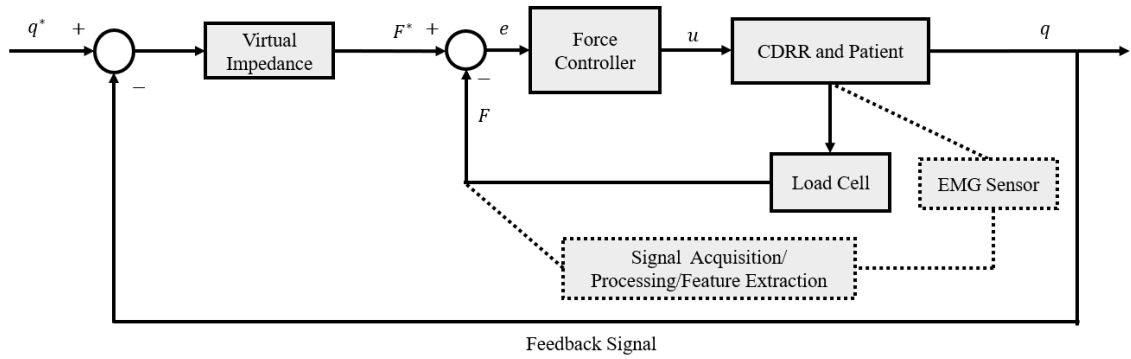

(a) Force/EMG based Impedance control strategy

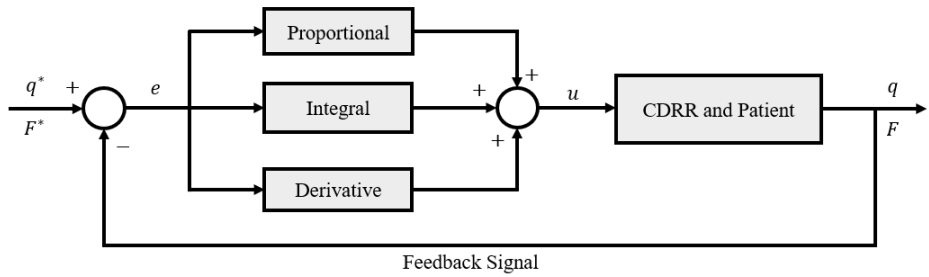

(b) Simple PID control strategy

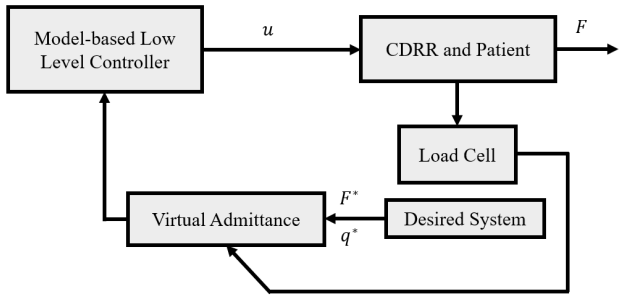

(c) Admittance control strategy

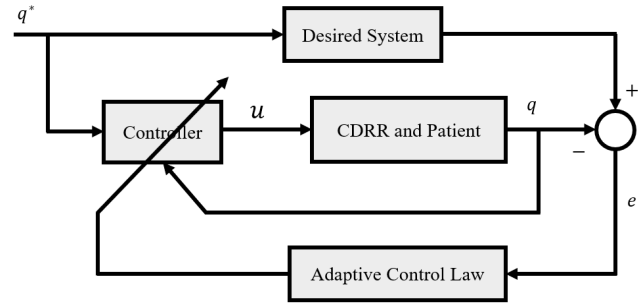

(d) Adaptive control strategy

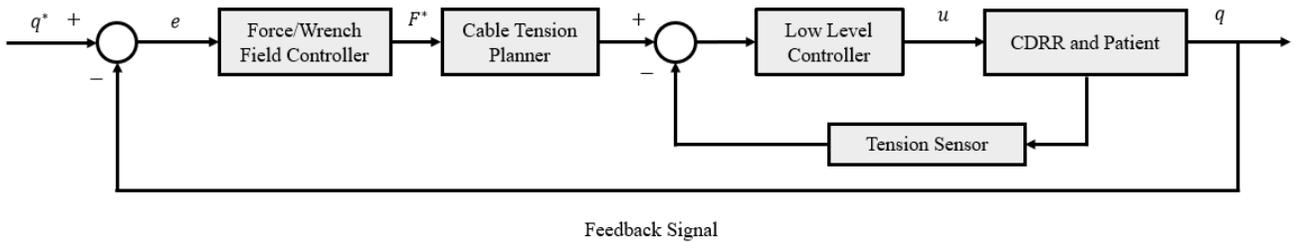

(e) Assist-as-needed control strategy

FIGURE 4: Control strategies of cable driven rehabilitation robot: $q^*$ means the referenced trajectory, $q$ means the actual motion trajectory, $F^*$ means the referenced force trajectory, $F$ means the actual force trajectory, Error $e$ is the deviation of the actual trajectory from the reference trajectory.

virtual admittance block updates patient's desired trajectory based on the interaction force between CDRR and patient, measured by load cell. The CDRRs used admittance based control are summarized in Table 11.

**Upper and Lower Limbs:** admittance-based controllers are found in few CDRRs and their performance was verified through experimental investigation and simulation, both for upper limb rehabilitation, such as Planar Cable-driven Parallel Robot [51], [52] and PACER robot [53], [54] and lower limb rehabilitation, such as Underactuated Lower Limb Exoskeleton [171] and LOKOIRAN [105].

### D. ADAPTIVE CONTROL STRATEGY

Adaptive control strategies offer better performance in the presence of exogenous disturbances, imperfect modeling, adverse system conditions, structural damage, changes in system dynamics, and cable or/and actuator failure [186]. Although other control strategies with fixed gains could also deal with adverse system conditions, but accurate modeling of the system and knowledge of uncertainty bounds are required. Adaptive control strategy has the capabilities for automatic tuning of controllers, and maintaining the system performance when the system parameters vary in unforeseen conditions, without excessively relying on system modeling.

Fig. 4 (d) presents the overall architecture of an adaptive controller with four main parts. It is assumed that the system model is known but it consists of uncertain parameters. The desired system block provides a reference behavior. The CDRR and patient block represents the actual dynamical system. The difference between desired and actual trajectories is then fed to the controller, in which the control law is parameterized with a number of adjustable parameters. The role of adaptive control law block is to update the controller parameters based on the performance error signal. The CDRRs used adaptive controller are summarized in Table 11.

**Upper and lower limb:** to compensate the model uncertainties and external disturbance of upper limb CDRRs, an ILC adaptive control is used in CARR-4 [135], and an adaptive fuzzy sliding mode control is employed in Robotic Exoskeleton by [144]. The use of adaptive controllers is also studied in CDRRs for lower limb such as Soft Exosuit; hip and ankle [165], Sophia-3 and Sophia-4 [90].

### E. ASSIST-AS-NEEDED (AAN) BASED CONTROL STRATEGY

An Assist-as-needed control reacts to trajectory error of robot end-effector and provides the patient with minimal amount of assistance necessary to complete a movement, while encouraging patient to make significant effort. This type of control strategy is used in many CDRRs, [8], [9], [64], [74], [155]. Fig. 4 (e) shows the overall block diagram of an Assist-as-needed control system.

Force field and wrench field are two types of AAN control strategy used in CDRRs. In the former strategy, a force field is generated around the referenced trajectory, which is then deployed to plan the cables tension. The low-level controller takes the tension commands and maintain the cables in tension. In the latter strategy, a six-DoFs wrench field is constructed based on the deviation from the desired pose to generate dexterous manipulation, including the three rotation and three translation. Compare to the force field Assist-as-needed controller, wrench field type provides the both force and torque commands to train and assist the patient. The CDRRs used AAN are summarized in Table 11.

**Upper and lower limb:** the performance of a force field Assist-as-needed control when patient pulls the end-effector of an anthropomorphic arm is studied in [8]. Similar control strategy is used to perform real-time estimation of glenohumeral joints rotation [187]. A six-DoFs wrench field AAN controller is deployed in CAREX-7 [9], [74]. The research on AAN control is reported in several other upper limb applications, such as C-ALEX [98], CDULRR [155], and CDWRR [64], and lower limb rehabilitation by TPAD [110].

### F. OTHER CONTROL METHODS

There are some other control strategies that yet have received less attention in designing and developing CDRRs.

**Sliding mode control strategy** is used as a robust control strategy for nonlinear systems, in which the tracking error approaches to zero asymptotically in the presence of nonlinear disturbances, [149], [150]. For instance, it is deployed for controlling damper like joints of BJS [63]. The effectiveness of the controller is validated through clinical trial.

**Iterative learning control strategy** can be referred as a trajectory tracking control strategy, which works in a repetitive mode. The robustness of iterative learning control against external disturbance and uncertainties in CDRRs is studied in [135]. The performance of this controller is evaluated through simulation and improvement in tracking performance was observed. Another CDRR is developed with this control strategy via using error information of the two adjacent iteration in Hand exoskeleton robot [128] and its performance was validated through the tracking results.

**Resistance compensation control strategy** was proposed to avoid undesired resistance in CDRRs [127]. This control strategy greatly solve the friction issue in CDRRs through resistance or friction compensation.

**Hierarchical cascade control strategy** consists of three levels of high, mid and low level controllers is to perform assistance level estimation, adaptive gap compensation, and adaptive position control with friction compensation, respectively. A soft exoskeleton robot for upper limb rehabilitation is reported in literature with this type of controller [117].

**Intelligent stretching control** is suitable for specific tasks, such as passive range of motion test during diagnosis and the passive stretching treatment. In an experiment, the patients were asked to relax, while Intelliarm [86] stretch the patient's joints with a specific commanded velocity.

**Patient-cooperative** and **Intention-driven control strategy**, are two other types of controller to support and encourage patients participate actively in the rehabilitation, which

TABLE 10: Controller details

| CDRR's Name; Reference | Control's level | Modeling approach | Performance | Operating space | Stability |
|---|---|---|---|---|---|
| **Impedance control strategy** | | | | | |
| Wearable Soft Orthotic Device [58] | Low-level | Model-free | Angular and linear misalignments error | Configuration space | N/A |
| NEUROExos [61] | High-level | Model-based | Steady-state angular error, gravity compensation | Configuration space | Asymptotic |
| ATD [55] | High-level | Model-based | Force tracking error, gravity compensation | Task space | Asymptotic |
| SNU Exo-Glove [124] | Low-level | Model-based | Cartesian error for task space tracking | Configuration space | N/A |
| L-EXOS [87] | High-level | Model-based | Cartesian error for task space tracking, gravity compensation | Task space | N/A |
| X-Arm-2 [141] | Mid-level | Model-based | Friction compensation | Configuration space | Asymptotic |
| CADEN 7 [143] | High-level | Model-based | Joints trajectory tracking error | Configuration space | N/A |
| SUEFUL-7 [72] | Low-level | Hybrid | Joints trajectory tracking error | Configuration space | N/A |
| Soft-Actuated Exoskeleton [146] | Low-level | Model based | Joints trajectory tracking error | Task space | N/A |
| Upper-Limb Powered [147] | High-level | Model based | Joints trajectory tracking error | Configuration space | N/A |
| Sophia-3 and Sophia-4 [90] | High+Low-level | Hybrid | Cartesian error for task space tracking | Task space | N/A |
| Powered Ankle Prostheses [102] | High-level | Model based | Cartesian error for task space tracking | Task space | N/A |
| LOPES [95] [96] [97] | High-level | Model based | Joint angle trajectory tracking error, cartesian impedance error, external disturbance rejection, gravity compensation | Task space | N/A |
| XoR [167] | Low-level | Model based | Force tracking error, external disturbance rejection, gravity compensation | N/A | N/A |
| Lower limb Rehab. Robot [100] | High-level | Model based | Cartesian error for task space tracking | Task space | N/A |
| **Proportional integral derivative based control strategy** | | | | | |
| Elbow Exoskeleton [85] | Mid-level | Model-based | Joint angle tracking error | Task space | Asymptotic |
| Wearable Rehab. Robo. Hand [84] | High-level | Model-free | Joint angle tracking error | Task space | N/A |
| Tendon Driven Hand orthosis [66] | Low-level | Model-free | Joint angles tracking of the index finger | Configuration space | N/A |
| ABLE [134] | Low-level | Model-free | Cartesian error for task space tracking, gravity compensation | Task space | N/A |
| 9-DoFs Rehab. Robot [79], [80] | Mid-level | Model-based | Position, force tracking error, friction, gravity compensation, | Configuration space | N/A |
| Upper Limb Rehab. Robot [138] | High+low-level | Model-based | Trajectory tracking error, friction, gravity compensation, | Task space | Stability-Lyapunov |
| ULERD [71], [75] | Low-level | Model-free | Joint angel traj. tracking error | Task space | Asymptotic |
| MULOS [89] | Low-level | Model-free | Quantitative measurements of error | N/A | N/A |
| Mirror-Image Motion Dev. [156] | Low-level | Model-free | Quantitative measurements of error | Configuration space | N/A |
| CDPR [113] | High+low-level | Model-based | Cartesian error for task space tracking, external disturbance rejection | Configuration and task space | Stability-Lyapunov |
| CDRR [158] | Low-level | Model-free | Position tracking error | N/A | N/A |
| Polycentric ankle exoskeleton [164] | Low-level | Model free | Joint angle tracking error | Configuration space | N/A |
| Powered Lower Limb Orthosis [166] | Low-level | Model-free | | N/A | Asymptotic |

Controller details of CDRRs

TABLE 11: Controller details

| CDRR's Name; Reference | Control's level | Modeling approach | Performance | Operating space | Stability |
|---|---|---|---|---|---|
| **Admittance based control strategy** | | | | | |
| Planar Cable-Driven Robot [51], [52] | Low-level | Hybrid | Trajectory tracking error in task space | Task space | N/A |
| PACER [53], [54] | High-level | Model based | Position and orientation error for task space tracking | Task space | N/A |
| Underactuated Lower Limb Exoskeleton [171] | High+low-level | Model based | Power augmentation error in load-carrying | Configuration space | N/A |
| LOKOIRAN [105] | Low-level | Model free | Friction and Gravity Compensation | Configuration space | N/A |
| **Adaptive control strategy** | | | | | |
| CARR-4 [135] | High +low -level | Model-free control | Position tracking errors | Configuration space | N/A |
| Robotic exoskeleton [144] | Mid -level | Model-free control | Cartesian error for task space tracking | Configuration and task space | Stability-Lyapunov |
| Sophia-3 and Sophia-4 [90] | High +low -level | Model-free control | Cartesian error for task space tracking | Task space | N/A |
| Soft Exosuit [157] | Low -level | Model-free control | RMS error for the peak force | Task space | N/A |
| **Assist-as-needed based control strategy** | | | | | |
| CDWRR [64] | Mid-level | Model-based | Joint angle tracking results | Configuration space | N/A |
| CAREX [8] | High+low-level | Model-based | Cartesian error for task space tracking, friction and gravity compensation | Task space | N/A |
| CAREX-7 [9], [74] | High+low-level | Model-based | Cartesian error for task space tracking, friction and gravity compensation | Configuration and task space | Stability-Lyapunov |
| CDULRR [155] | Mid-level | Model-based | Cartesian error for task space tracking, gravity compensation, external disturbances rejection | Configuration and task space | N/A |
| C-ALEX [98] | High+low-level | Model-based | Cartesian error for tracking, friction & gravity compensate | Task space | N/A |
| TPAD [110] | High+low-level | Model-based | Cartesian error for task space tracking, friction compensation | Configuration and task space | N/A |

have been employed in HIT-Glove [88] and Upper Limb Exoskeleton [133], respectively.

### G. DISCUSSIONS ON CONTROL STRATEGIES

The control strategies for CDRRs should consider the system coupling (robot-human interaction) in the control loop. A sole force or position controller could not ensure safe dynamic interaction between robot and human [188]. The pros and cons of mostly used control strategies in CDRRs are discussed here and also summarized in Table 12.

The impedance control strategy is largely used in CDRRs, as it provides stable interaction between human and robot. The impedance value or desired interaction force is selected by the therapists based on their experience and considering the level of patients' disability. The challenge is to ensure an intrinsically stiff actuation, which can be achieved by compensating the natural impedance of system, and other effects due to the friction and gravity [189]. Impedance control allows the actuation in three modes, robot in charge, patient in charge, and therapist in charge. In the robot in charge mode, the value of impedance is chosen to be high, and robot guides the patient to follow the desired trajectory. In the patient in charge mode, the value of impedance is set to be low, and the robot does not hinder the motion by the patient. In the therapist in charge mode, the value of impedance is chosen manually to provide a corrective or supportive torques for patient training [96].

The impedance control strategy allows the real-time adjustment of impedance parameters such as damping, inertia, and stiffness, without leading to any significant stability problems [190]. Impedance control can be considered as a suitable candidate for interaction control strategy as it does not relay on the precise knowledge of external parameters [191]. Unlike other control strategies, where the force and position are controlled separately, impedance controller controls the dynamic relationship between the force and velocity, also known as impedance of the CDRR's end-effector [192], [193]. The impedance control does not have significant difference with position control except it offers less interaction torque and large spatial variety [190]. Its accuracy can be further improved by using low-friction joints, inner loop torque sensors or direct drives. This control strategy also circumvents the need of using force sensors, which are generally expensive and sensitive to temperature changes.

EMG-based impedance controller is a control strategy, which uses the patient's electromyography biological signals to control the robot. It is suitable where muscle strength needs to be monitored and improved. A potential drawback of EMG based controller, is the lack of patient in charge mode or the voluntary movements of patient, as the change in muscles strength trigger the controller [189]. Another problem is the difficulty in precise generation of the force base on the varying EMG signals in real time. This problem can be addressed in future research by using a combination of multiple biological signals such as electroencephalogram, electromyogram, and electrooculogram.

Admittance based control relies on measuring the force exerted by the patient to generate the corresponding displacement of the assistive robot. This control strategy provides higher accuracy for contact-free tasks, but dynamic interactions between human and robot can lead to unstable performance [22]. While both impedance and admittance controllers could be used for interaction with soft environments, their performance are opposite. The performance of admittance control declines with the increase in the environment's stiffness. In contrast, the performance of impedance control decline with the decrease in the environment's stiffness [194]. Additionally, the performance of impedance controller depends on the precision of the position sensors, whereas the performance of admittance controller depends on the accuracy of the force/torque sensors. Admittance control may be preferred, where the accurate information of interaction force between patient and robot is required [27]. A combined use of impedance and admittance control in CDRRs could benefit from the advantages of both methods.

Adaptive control is preferred when dealing with uncertain dynamic systems in CDRRs. This type of control does not rely on the modelling accuracy, and has the ability to improve its performance under adverse and unforeseen conditions. It can deal with uncertainty in both kinematic and dynamics of system being controlled. Adaptive control strategy is very effective to assist the patient in recovering the balance, especially when the stability problem is detected, and It can also compensate the friction and hysteresis backlash [195]. Adaptive control shows a better tracking performance compare to the simple PD controller, with same feedback gain. Instead of using high controller gains, adaptive control learns from the system dynamics, patient's effort and ability [196]. However, it is rarely optimal as it does not consider or rely on the accurate modelling of the rehabilitation robot [197].

The Assist-As-Needed is one of the most important control strategies. AAN parameters are adjusted according to the required level of assistance, which not only varies from patient to patient but also depends on different rehabilitation exercises [198]. The Assist-As-Needed control systems are evaluated for CDRRs using both lab experiments and clinical investigation. A potential drawback of assist-as-needed based controller is the reliance on the accurate knowledge of patient's functional ability. An inconsistent or inaccurate estimation of patient's functional movement can lead to failure of this approach [199]. Therefore, a challenge is how to appropriately estimate the patient's functional ability to precisely generate required level of robotic assistance with this type of controller [200]. Less work is done in literature on precise estimation of patient's functional ability [201]. These issues can be further analysed through clinical trials and considered as a direction of future work.

Direct comparative experiments on different control strategies of CDRRs are very limited in literature. Stein et al. [202] is one of the very few works with comparative experimental study between the Impedance and Resistance-based control strategies and reported no significant difference in performance. In another research, Lance et al. [203] claimed comparing two assist-as-needed control strategies, one with interlimb coordination constraints that showed more pronounced and fast recovery. Based on our knowledge, there is no other direct comparison of control strategies in CDRRs. The control strategies are mostly evaluated in different experimental environments, with different cable driven rehabilitation robots, and for different target movements for rehabilitation. The main pros and cons of control strategies based on the available literature are summarized in Table 12.

## V. DISCUSSION AND FUTURE CHALLENGES

There is a growing interest in research and development of CDRRs for rehabilitation. Around 200 references and research articles on CDRRs are found and reviewed in this work for lower limb, upper limb and waist rehabilitation.

Concerning different training applications, rehabilitation robots should meet different requirements such as: a) repeated facilitation training, b) uniform performance in a long period of time, c) safe interaction with patients, d) back-drivability for patient-in-charge mode, e) quantitative measures for performance analysis, and f) adaptability and reconfigurability for different patients.

A cable-driven actuation can help meeting the above requirements. In CDRRs, actuators are placed at the base and joints of CDRR are actuated through cables, which make the CDRR very light weight. So, they can perform repeated training with less energy. Their light weight also provides the space for more sensors to quantitatively measure the performance of CDRRs. As the actuators are not attached to the joints, they also have less inertia, which not only increase the safety but also help in obtaining a uniform performance with CDRRs. Additionally, the cable driven actuators are back drivable and by changing cable length they provide size adaptability for different patients.

There are several key technologies reported in literature that successfully used cable driven actuation for rehabilitation applications, such as CAREX-7 [9], [74], Tendon Driven Hand orthosis [66], Soft Exosuit: ankle [108], C-

TABLE 12: Advantages and disadvantages of mostly used control strategies in CDRRs

| Control strategies | Advantages | Disadvantages |
|---|---|---|
| Impedance based control strategy | <ul><li>stable interaction between human and robot.</li><li>allows adjustment of impedance value based on the therapist's experience and patient's disability.</li><li>control ability in three modes, robot in charge, patient in charge, and therapist in charge.</li><li>real-time adjustment of impedance parameters such as damping, inertia, and stiffness.</li><li>does not relay on the precise knowledge of external parameters.</li><li>monitor and improve the muscle strength by using EMG-based impedance control.</li></ul> | <ul><li>performance declines with the decrease in the environment's stiffness.</li><li>requires the force sensors, which are generally expensive and sensitive to temperature changes.</li><li>does not allow the voluntary movements of patient, as the change in muscles strength trigger the controller (EMG-base impedance control).</li><li>faces difficulty in precisely generation the force base on the varying EMG signals in real time (EMG -base impedance control).</li><li>faces difficulty in ensuring the intrinsically stiff actuation system.</li></ul> |
| Proportional integral derivative based control strategy | <ul><li>easy to apply for CDRRs.</li><li>robust performance with simple structure.</li><li>robust to tuning mismatches.</li><li>asymptotic convergence of errors in task space.</li><li>independent control of the joints in joint space.</li></ul> | <ul><li>a sole force or position PID controller could not ensure safe dynamic interaction between robots and human.</li><li>produces large overshoots which are not desirable for CDRRs.</li></ul> |
| Admittance based control strategy | <ul><li>provides higher accuracy for contact-free tasks.</li><li>works well when accurate information of interaction force between patient and robot is available.</li></ul> | <ul><li>dynamic interactions between human and robot can lead to unstable performance.</li><li>performance declines with the increase in the environment's stiffness.</li></ul> |
| Adaptive control strategy | <ul><li>robust to uncertain dynamic systems.</li><li>ability to improve its performance under adverse and unforeseen conditions.</li><li>suitable if system stability problem is detected.</li><li>compensates the friction and hysteresis backlash.</li></ul> | <ul><li>it is rarely optimal as it does not rely on the accurate modelling of the rehabilitation robot.</li></ul> |
| Assist-as-needed control strategy | <ul><li>parameters are adjusted based on the level of assistance required by the patient or rehabilitation exercise.</li><li>evaluated for CDRRs using both lab experiments and clinical investigation.</li></ul> | <ul><li>inconsistent and inaccurate estimation of patient's functional movement can be a major limitation in use of this approach</li><li>generating accurate level of robotic assistance in real-time based on estimation of the patients' functional ability is challenging</li></ul> |

ALEX [98], CAREX [8], NeReBot [91], [151], [152], LOPES [95]–[97], and String Man [106]. These CDRRs offer low inertia, light weight, high payload-to-weight ratio, large work space, and adaptability. CAREX-7 [9], [74] is a 7-DOF CDRR that is designed for the movement training of the upper limb. The mechanical structure of CAREX-7 is light weight and have less moving inertia, as it consists of four light weight cuffs and eight cables. Tendon Driven Hand orthosis [66] is a low weight design and it also ensure safety because it does not require custom joint alignment and has low inertia. Soft Exosuit: ankle [108] is a gait rehabilitation robot and it was proposed to assist ankle plantarflexion and dorsiflexion. This allows the actuators to be placed for away from moving joints, which lead to large workspace. C-ALEX [98] is a lower limb CDRR with high payload-to-weight ratio. It consists of few cuffs and cables and designed for gait rehabilitation. These works not only prove the effectiveness through experimental investigations but also demonstrated the satisfactory performance in the clinical study.

Existing cable driven rehabilitation robots have demonstrated various capabilities to assist and train patients and have shown effectiveness in rehabilitation. Although CDRRs have a lot of promising features, there are also some limitations and deficiencies that are summarized as follows:

- The CDRRs can exert only tensile forces, as the cables of CDRRs can only be used to pull (and not to push).
- The cables of CDRRs must be in tension at all the time, making it more difficult to optimise the effective workspace of the rehabilitation robot.
- The elasticity or flexibility of cables of CDRRs causes undesirable vibrations, which may contribute to having position and orientation errors.

- The CDRRs have high maintenance requirements mainly due to the cable breakage, slackening, and friction of the system.

These limitations increase the complexity of kinematic and dynamic modelling and decrease the stiffness of CDRRs. It is also challenging to design and deploy robust controllers for CDRRs that can deal with flexibility, friction, and vibrations of the cables. A few key research and solutions focusing on these problems include; friction compensation [204], self healing concept [205], layer jamming [206], and singular perturbation approach based modeling [18]. A novel friction-tension mechanic model was proposed by Youngsu et al. [204] to compensate the friction between the pulleys and cables. In which two possible types of cable-pulley transmission are considered, i.e, free-free and fixed-free ended. To solve the stiffness issue with cables, a variable stiffness cable with self-healing capability was introduced by Alice et al. [205] that changes its stiffness relative to the variation in temperature. Another research work on cable stiffness was presented by the Yong et al. [206] by using layer jamming concept, in which cable's property changes from flexible to highly stiff depending on if vacuum is applied or not. To deal with the flexibility issue of cables, a dynamic modeling and control strategy based on singular perturbation approach was introduced by [18]. However, these concepts still need to be implemented and verified with CDRRs and remains as continuing research area.

The evidence from this review suggests that future work and scientific research on CDRRs could consider the following aspects and areas;

- The existing CDRRs for rehabilitation of multiple human joints consists of many passive and active cables, which make the structure of CDRRs complex, heavy and bulky. Future CDRRs should be more compact, lightweight and portable for rehabilitation of elderly people at home and assisting patients with limited access to rehabilitation centres. More research is needed to address challenges of designing CDRRs for simultaneous rehabilitation of multiple human joints.
- The cost of developing and verifying CDRRs, and need for expensive sensors such as force/torque, tension, and EEG sensors, and costly clinical trials are major hindrance in extending the application of CDRRs. Addressing these problems may dramatically impact on the use of CDRRs.
- The few shortcomings in the performance of CDRRs that need more attention in future research. For instance, compare to serial type CDRRs, the parallel type ones normally have better adaptability, but they have to deal with problems such as uncertainty, controllability and limited workspace.
- The control strategy plays a vital role in effective performance of CDRRs, where there is still room for improvement. Five main categories of control strategies in CDRRs are reviewed in this work, but only two of them have gone through clinical investigation. Moreover, the use of bio-signals in control strategies has attracted less attention. The future research could focus on clinical verification of control strategies, integration of more bio-signals such as EEG and EOG in the control systems, and human-CDRR interactive control based on reinforcement learning. Additionally, more research is needed on optimizing control strategies for rehabilitation in terms of performance and rate of recovery.
- The human-machine interface allows effective incorporation of CDRRs and motivate the patients toward rehabilitation. More work is needed for developing dexterous manipulation of CDRRs with HMI, and studying physical and clinical implications of long-term use of HMIs in rehabilitation exercise. Recently, the entertainment gaming industry have developed several new interesting VR-based interfaces to capture the motion of the user. In future, these interfaces can be integrated in CDRRs to facilitate the rehabilitation training.
- Finally, the clinical verification of the CDRRs has received very limited attention. Less than 5% of CDRRs in this review have gone through clinical investigations. These clinical studies are at initial level and there is no set of widely accepted clinical criteria for CDRRs. The updated or advanced versions of CDRRs are not proposed to overcome the shortcoming based on the clinical study. The future research should focus on large-scale clinical investigation of CDRRs. Furthermore, clinical investigations need to provide the clear performance comparison over the manually assisted training.

## VI. CONCLUSION

Rehabilitation robots including cable-driven types, have gained a lot of attention to cope with high demands of physio therapy and reduce reliance on professional therapists. Around 200 references and research articles on CDRRs are reviewed in this work with the aim of understanding the successes and shortfalls of existing developments and future needs. To facilitate the discussion, CDRRs are categorized into three major categorizes for upper limb, lower limb, and waist rehabilitation. For each group of robots, target movements of rehabilitation are identified and existing CDRRs are reviewed in terms of type of actuators, sensors, controllers, physical interactions with patients, and human-machine interface.

Exisiting CDRRs offer significant advantages in terms of low inertia, light weight, high payload-to-weight ratio, large work space and configurability. They enable treatments with high intensity, re-peatability and real time measurement of patient performance. They can be used independently or collaboratively to support therapist in the rehabilitation process. Various control strategies are successfully developed for CDRRs, which are mainly categorized under Impedance-based, PID-based, Admittance-based, Assist-as-needed (AAN) and Adaptive controllers. The clinical studies on CDRRs performance in rehabilitation settings are also promising, though there is room for improvement.

Future works can focus on designing more compact and portable CDRRs, as well as addressing the challenges of developing CDRRs for simultaneous rehabilitation of multiple human joints. Bio-signals are invaluable source of data for the purpose of controlling CDRRs. Further enhancement in the performance and speed of control strategies is needed to deal with uncertainties of physical interaction between human and CDRRs. More attention in research is also required on clinical study and verification of CDRRs in clinical settings.

## REFERENCES

[1] "Disability and health," https://www.who.int/news-room/fact-sheets/detail/disability-and-health, accessed: Sept. 09, 2019. 1

[2] "World report on disability," https://www.who.int/disabilities/world_report/2011/report.pdf, accessed: Sept. 09, 2019. 1

[3] S. Barreca, S. L. Wolf, S. Fasoli, and R. Bohannon, "Treatment interventions for the paretic upper limb of stroke survivors: a critical review," Neurorehabilitation and neural repair, vol. 17, no. 4, pp. 220–226, 2003. 1

[4] B. H. Dobkin, "Strategies for stroke rehabilitation," The Lancet Neurology, vol. 3, no. 9, pp. 528–536, 2004. 1

[5] S. H. Lee, G. Park, D. Y. Cho, H. Y. Kim, J.-Y. Lee, S. Kim, S.-B. Park, and J.-H. Shin, "Comparisons between end-effector and exoskeleton rehabilitation robots regarding upper extremity function among chronic stroke patients with moderate-to-severe upper limb impairment," Scientific reports, vol. 10, no. 1, pp. 1–8, 2020. 1

[6] L. Dovat, O. Lambercy, R. Gassert, T. Maeder, T. Milner, T. C. Leong, and E. Burdet, "Handcare: a cable-actuated rehabilitation system to train hand function after stroke," IEEE Transactions on Neural Systems and Rehabilitation Engineering, vol. 16, no. 6, pp. 582–591, 2008. 1

[7] M. Shoaib, C. Y. Lai, and A. Bab-Hadiashar, "A novel design of cable-driven neck rehabilitation robot (carneck)," in 2019 IEEE/ASME International Conference on Advanced Intelligent Mechatronics (AIM). IEEE, 2019, pp. 819–825. 1

[8] Y. Mao and S. K. Agrawal, "Design of a cable-driven arm exoskeleton (carex) for neural rehabilitation," IEEE Transactions on Robotics, vol. 28, no. 4, pp. 922–931, 2012. 1, 4, 5, 6, 9, 15, 17, 19

[9] X. Cui, W. Chen, X. Jin, and S. K. Agrawal, "Design of a 7-dof cable-driven arm exoskeleton (carex-7) and a controller for dexterous motion training or assistance," IEEE/ASME Transactions on Mechatronics, vol. 22, no. 1, pp. 161–172, 2016. 1, 4, 10, 15, 17, 18, 19

[10] A. M. Dollar and H. Herr, "Lower extremity exoskeletons and active orthoses: challenges and state-of-the-art," IEEE Transactions on robotics, vol. 24, no. 1, pp. 144–158, 2008. 1

[11] S. J. Ball, I. E. Brown, and S. H. Scott, "A planar 3dof robotic exoskeleton for rehabilitation and assessment," in 2007 29th Annual International Conference of the IEEE Engineering in Medicine and Biology Society. IEEE, 2007, pp. 4024–4027. 1, 4, 5, 6, 8, 9

[12] R. Gopura, D. Bandara, K. Kiguchi, and G. K. Mann, "Developments in hardware systems of active upper-limb exoskeleton robots: A review," Robotics and Autonomous Systems, vol. 75, pp. 203–220, 2016. 1

[13] H. Krebs, , and B. Volpe, "Rehabilitation robotics," Handbook of clinical neurology, vol. 110, pp. 283–294, 2013. 1

[14] R. Colombo and V. Sanguineti, "Rehabilitation robotics: Technology and application," 2018. 1

[15] Z. Z. Bien and D. Stefanov, Advances in rehabilitation robotics: Human-friendly technologies on movement assistance and restoration for people with disabilities. Springer Science & Business, 2004, vol. 306. 1

[16] E. Rocon and J. L. Pons, Exoskeletons in rehabilitation robotics: Tremor suppression. Springer, 2011, vol. 69. 1

[17] T. Bruckmann and A. Pott, Cable-driven parallel robots. Springer, 2012, vol. 12. 2

[18] M. A. Khosravi and H. D. Taghirad, "Dynamic modeling and control of parallel robots with elastic cables: singular perturbation approach," IEEE Transactions on Robotics, vol. 30, no. 3, pp. 694–704, 2014. 2, 20

[19] S. Kawamura, H. Kino, and C. Won, "High-speed manipulation by using parallel wire-driven robots," Robotica, vol. 18, no. 1, pp. 13–21, 2000. 2

[20] M. Shoaib, J. Cheong, D. Park, and C. Park, "Composite controller for antagonistic tendon driven joints with elastic tendons and its experimental verification," IEEE Access, vol. 6, pp. 5215–5226, 2017. 2

[21] M. Babaiasl, S. H. Mahdioun, P. Jaryani, and M. Yazdani, "A review of technological and clinical aspects of robot-aided rehabilitation of upper-extremity after stroke," Disability and Rehabilitation: Assistive Technology, vol. 11, no. 4, pp. 263–280, 2016. 2

[22] P. Maciejasz, J. Eschweiler, K. Gerlach-Hahn, A. Jansen-Troy, and S. Leonhardt, "A survey on robotic devices for upper limb rehabilitation," Journal of neuroengineering and rehabilitation, vol. 11, no. 1, p. 3, 2014. 2, 18

[23] A. S. Niyetkaliyev, S. Hussain, M. H. Ghayesh, and G. Alici, "Review on design and control aspects of robotic shoulder rehabilitation orthoses," IEEE Transactions on Human-Machine Systems, vol. 47, no. 6, pp. 1134–1145, 2017. 2

[24] H. S. Lo and S. Q. Xie, "Exoskeleton robots for upper-limb rehabilitation: State of the art and future prospects," Medical engineering & physics, vol. 34, no. 3, pp. 261–268, 2012. 2

[25] A. Basteris, S. M. Nijenhuis, A. H. Stienen, J. H. Buurke, G. B. Prange, and F. Amirabdollahian, "Training modalities in robot-mediated upper limb rehabilitation in stroke: a framework for classification based on a systematic review," Journal of neuroengineering and rehabilitation, vol. 11, no. 1, p. 111, 2014. 2

[26] R. C. Loureiro, W. S. Harwin, K. Nagai, and M. Johnson, "Advances in upper limb stroke rehabilitation: a technology push," Medical & biological engineering & computing, vol. 49, no. 10, p. 1103, 2011. 2

[27] T. Proietti, V. Crocher, A. Roby-Brami, and N. Jarrasse, "Upper-limb robotic exoskeletons for neurorehabilitation: a review on control strategies," IEEE reviews in biomedical engineering, vol. 9, pp. 4–14, 2016. 2, 18

[28] S. Masiero, M. Armani, G. Rosati et al., "Upper-limb robot-assisted therapy in rehabilitation of acute stroke patients: focused review and results of new randomized controlled trial," J Rehabil Res Dev, vol. 48, no. 4, pp. 355–366, 2011. 2

[29] N. Norouzi-Gheidari, P. S. Archambault, and J. Fung, "Effects of robot-assisted therapy on stroke rehabilitation in upper limbs: systematic review and meta-analysis of the literature." Journal of Rehabilitation Research & Development, vol. 49, no. 4, 2012. 2

[30] R. Morales, F. J. Badesa, N. García-Aracil, J. M. Sabater, and C. Pérez-Vidal, "Pneumatic robotic systems for upper limb rehabilitation," Medical & biological engineering & computing, vol. 49, no. 10, p. 1145, 2011. 2

[31] N. Jarrassé, T. Proietti, V. Crocher, J. Robertson, A. Sahbani, G. Morel, and A. Roby-Brami, "Robotic exoskeletons: a perspective for the rehabilitation of arm coordination in stroke patients," Frontiers in human neuroscience, vol. 8, p. 947, 2014. 2

[32] P. K. Jamwal, S. Hussain, and S. Q. Xie, "Review on design and control aspects of ankle rehabilitation robots," Disability and Rehabilitation: Assistive Technology, vol. 10, no. 2, pp. 93–101, 2015. 2

[33] G. Chen, C. K. Chan, Z. Guo, and H. Yu, "A review of lower extremity assistive robotic exoskeletons in rehabilitation therapy," Critical Reviews™ in Biomedical Engineering, vol. 41, no. 4-5, 2013. 2

[34] T. Yan, M. Cempini, C. M. Oddo, and N. Vitiello, "Review of assistive strategies in powered lower-limb orthoses and exoskeletons," Robotics and Autonomous Systems, vol. 64, pp. 120–136, 2015. 2

[35] A. Pennycott, D. Wyss, H. Vallery, V. Klamroth-Marganska, and R. Riener, "Towards more effective robotic gait training for stroke rehabilitation: a review," Journal of neuroengineering and rehabilitation, vol. 9, no. 1, pp. 1–13, 2012. 2

[36] J.-M. Belda-Lois, S. Mena-del Horno, I. Bermejo-Bosch, J. C. Moreno, J. L. Pons, D. Farina, M. Iosa, M. Molinari, F. Tamburella, A. Ramos et al., "Rehabilitation of gait after stroke: a review towards a top-down approach," Journal of neuroengineering and rehabilitation, vol. 8, no. 1, p. 66, 2011. 2

[37] I. Díaz, J. J. Gil, and E. Sánchez, "Lower-limb robotic rehabilitation: literature review and challenges," Journal of Robotics, vol. 2011, 2011. 2

[38] S. Hussain, "State-of-the-art robotic gait rehabilitation orthoses: design and control aspects," NeuroRehabilitation, vol. 35, no. 4, pp. 701–709, 2014. 2

[39] W. Huo, S. Mohammed, J. C. Moreno, and Y. Amirat, "Lower limb wearable robots for assistance and rehabilitation: A state of the art," IEEE systems Journal, vol. 10, no. 3, pp. 1068–1081, 2014. 2

[40] P. Heo, G. M. Gu, S.-j. Lee, K. Rhee, and J. Kim, "Current hand exoskeleton technologies for rehabilitation and assistive engineering," In-

ternational Journal of Precision Engineering and Manufacturing, vol. 13, no. 5, pp. 807–824, 2012. 2

[41] P. S. Lum, S. B. Godfrey, E. B. Brokaw, R. J. Holley, and D. Nichols, "Robotic approaches for rehabilitation of hand function after stroke," American journal of physical medicine & rehabilitation, vol. 91, no. 11, pp. S242–S254, 2012. 2

[42] S. Balasubramanian, J. Klein, and E. Burdet, "Robot-assisted rehabilitation of hand function," Current opinion in neurology, vol. 23, no. 6, pp. 661–670, 2010. 2

[43] L. Marchal-Crespo and D. J. Reinkensmeyer, "Review of control strategies for robotic movement training after neurologic injury," Journal of neuroengineering and rehabilitation, vol. 6, no. 1, pp. 1–15, 2009. 2, 12

[44] H. Xiong and X. Diao, "A review of cable-driven rehabilitation devices," Disability and Rehabilitation: Assistive Technology, vol. 15, no. 8, pp. 885–897, 2020. 2

[45] Y. Cho, S. Muhammad, and J. Cheong, "Kinematic analysis of tendon driven robot mechanism for heavy weight handling," in 2015 12th International Conference on Ubiquitous Robots and Ambient Intelligence (URAI). IEEE, 2015, pp. 91–94. 2

[46] M. Shoaib, M. S. Khan, M. Asim, and J. Cheong, "Dynamic modeling and control of small scale wire driven robotic joints with wire elasticity," Journal of Nanoelectronics and Optoelectronics, vol. 13, no. 9, pp. 1389–1396, 2018. 2

[47] J. Cao, S. Q. Xie, R. Das, and G. L. Zhu, "Control strategies for effective robot assisted gait rehabilitation: the state of art and future prospects," Medical engineering & physics, vol. 36, no. 12, pp. 1555–1566, 2014. 2

[48] "https://www.progressiveautomations.com/pages/actuators." 3

[49] Y. Mao and S. K. Agrawal, "Transition from mechanical arm to human arm with carex: A cable driven arm exoskeleton (carex) for neural rehabilitation," in 2012 IEEE International Conference on Robotics and Automation. IEEE, 2012, pp. 2457–2462. 4, 5

[50] A. H. Stienen, E. E. Hekman, F. C. Van der Helm, G. B. Prange, M. J. Jannink, A. M. Aalsma, and H. Van der Kooij, "Dampace: dynamic force-coordination trainer for the upper extremities," in 2007 IEEE 10th International Conference on Rehabilitation Robotics. IEEE, 2007, pp. 820–826. 4, 5, 6, 9

[51] X. Jin, D. I. Jun, X. Jin, J. Seon, A. Pott, S. Park, J.-O. Park, and S. Y. Ko, "Upper limb rehabilitation using a planar cable-driven parallel robot with various rehabilitation strategies," in Cable-Driven Parallel Robots. Springer, 2015, pp. 307–321. 4, 6, 9, 15, 17

[52] X. Jin, D. I. Jun, X. Jin, S. Park, J.-O. Park, and S. Y. Ko, "Workspace analysis of upper limb for a planar cable-driven parallel robots toward upper limb rehabilitation," in 2014 14th International Conference on Control, Automation and Systems (ICCAS 2014). IEEE, 2014, pp. 352–356. 4, 5, 6, 9, 15, 17

[53] A. Alamdari and V. Krovi, "Parallel articulated-cable exercise robot (pacer): novel home-based cable-driven parallel platform robot for upper limb neuro-rehabilitation," in ASME 2015 International Design Engineering Technical Conferences and Computers and Information in Engineering Conference. American Society of Mechanical Engineers Digital Collection, 2016. 4, 6, 10, 15, 17

[54] ——, "Modeling and control of a novel home-based cable-driven parallel platform robot: Pacer," in 2015 IEEE/RSJ International Conference on Intelligent Robots and Systems (IROS). IEEE, 2015, pp. 6330–6335. 4, 5, 6, 10, 15, 17

[55] A. J. Westerveld, B. J. Aalderink, W. Hagedoorn, M. Buijze, A. C. Schouten, and H. van der Kooij, "A damper driven robotic end-point manipulator for functional rehabilitation exercises after stroke," IEEE transactions on biomedical engineering, vol. 61, no. 10, pp. 2646–2654, 2014. 4, 5, 6, 8, 13, 16

[56] S. J. Ball, I. Brown, and S. H. Scott, "Designing a robotic exoskeleton for shoulder complex rehabilitation," CMBES Proceedings, vol. 30, 2007. 4

[57] S. J. Ball, I. E. Brown, and S. H. Scott, "Medarm: a rehabilitation robot with 5dof at the shoulder complex," in 2007 IEEE/ASME international conference on Advanced intelligent mechatronics. IEEE, 2007, pp. 1–6. 4

[58] I. Galiana, F. L. Hammond, R. D. Howe, and M. B. Popovic, "Wearable soft robotic device for post-stroke shoulder rehabilitation: Identifying misalignments," in 2012 IEEE/RSJ International Conference on Intelligent Robots and Systems. IEEE, 2012, pp. 317–322. 4, 8, 16

[59] A. Niyetkaliyev, E. Sariyildiz, and G. Alici, "Kinematic modeling and analysis of a novel bio-inspired and cable-driven hybrid shoulder mechanism," Journal of Mechanisms and Robotics, vol. 13, no. 1, p. 011008, 2021. 4

[60] S. B. Kesner, L. Jentoft, F. L. Hammond, R. D. Howe, and M. Popovic, "Design considerations for an active soft orthotic system for shoulder rehabilitation," in 2011 Annual International Conference of the IEEE Engineering in Medicine and Biology Society. IEEE, 2011, pp. 8130–8134. 4, 8

[61] N. Vitiello, T. Lenzi, S. Roccella, S. M. M. De Rossi, E. Cattin, F. Giovacchini, F. Vecchi, and M. C. Carrozza, "Neuroexos: A powered elbow exoskeleton for physical rehabilitation," IEEE Transactions on Robotics, vol. 29, no. 1, pp. 220–235, 2012. 4, 8, 13, 16

[62] M. A. Laribi and M. Ceccarelli, "Design and experimental characterization of a cable-driven elbow assisting device," Journal of Medical Devices, 2021. 4

[63] C. Jarrett and A. McDaid, "Robust control of a cable-driven soft exoskeleton joint for intrinsic human-robot interaction," IEEE Transactions on Neural Systems and Rehabilitation Engineering, vol. 25, no. 7, pp. 976–986, 2017. 4, 8, 15

[64] W. Chen, X. Cui, J. Zhang, and J. Wang, "A cable-driven wrist robotic rehabilitator using a novel torque-field controller for human motion training," Review of Scientific Instruments, vol. 86, no. 6, p. 065109, 2015. 4, 8, 15, 17

[65] L. Yang, F. Zhang, J. Zhu, and Y. Fu, "A portable device for hand rehabilitation with force cognition: Design, interaction and experiment," IEEE Transactions on Cognitive and Developmental Systems, 2021. 4

[66] S. Park, L. Weber, L. Bishop, J. Stein, and M. Ciocarlie, "Design and development of effective transmission mechanisms on a tendon driven hand orthosis for stroke patients," in 2018 IEEE International Conference on Robotics and Automation (ICRA). IEEE, 2018, pp. 2281–2287. 4, 6, 8, 13, 16, 18, 19

[67] S. Min and S. Yi, "Development of cable-driven anthropomorphic robot hand," IEEE Robotics and Automation Letters, vol. 6, no. 2, pp. 1176–1183, 2021. 4

[68] K. Serbest, M. Kutlu, O. Eldogan, and I. Tekeoglu, "Development and control of a home-based training device for hand rehabilitation with a spring and cable driven mechanism," Biomedical Engineering/Biomedizinische Technik, 2021. 4

[69] J.-G. Yoon and M. C. Lee, "Design of tendon mechanism for soft wearable robotic hand and its fuzzy control using electromyogram sensor," Intelligent Service Robotics, pp. 1–10, 2021. 4

[70] P. M. Aubin, H. Sallum, C. Walsh, L. Stirling, and A. Correia, "A pediatric robotic thumb exoskeleton for at-home rehabilitation: the isolated orthosis for thumb actuation (iota)," in 2013 IEEE 13th International Conference on Rehabilitation Robotics (ICORR). IEEE, 2013, pp. 1–6. 4, 8

[71] Z. Song and S. Guo, "Design process of exoskeleton rehabilitation device and implementation of bilateral upper limb motor movement," Journal of Medical and Biological Engineering, vol. 32, no. 5, pp. 323–330, 2012. 4, 9, 13, 16

[72] R. A. R. C. Gopura, K. Kiguchi, and Y. Li, "Sueful-7: A 7dof upper-limb exoskeleton robot with muscle-model-oriented emg-based control," in 2009 IEEE/RSJ International Conference on Intelligent Robots and Systems. IEEE, 2009, pp. 1126–1131. 4, 10, 12, 13, 16

[73] F. Xiao, Y. Gao, Y. Wang, Y. Zhu, and J. Zhao, "Design of a wearable cable-driven upper limb exoskeleton based on epicyclic gear trains structure," Technology and Health Care, vol. 25, no. S1, pp. 3–11, 2017. 4, 9

[74] X. Cui, W. Chen, J. Zhang, and J. Wang, "Note: Model-based identification method of a cable-driven wearable device for arm rehabilitation," Review of Scientific Instruments, vol. 86, no. 9, p. 096107, 2015. 4, 10, 15, 17, 18, 19

[75] Z. Song, S. Guo, M. Pang, and S. Zhang, "Ulerd-based active training for upper limb rehabilitation," in 2012 IEEE International Conference on Mechatronics and Automation. IEEE, 2012, pp. 569–574. 4, 9, 13, 16

[76] F. Martinez, I. Retolaza, A. Pujana-Arrese, A. Cenitagoya, J. Basurko, and J. Landaluze, "Design of a five actuated dof upper limb exoskeleton oriented to workplace help," in 2008 2nd IEEE RAS & EMBS International Conference on Biomedical Robotics and Biomechatronics. IEEE, 2008, pp. 169–174. 4, 9

[77] F. Martinez, A. Pujana-Arrese, I. Retolaza, I. Sacristan, J. Basurko, and J. Landaluze, "Iko: a five actuated dof upper limb exoskeleton oriented to workplace assistance," Applied Bionics and Biomechanics, vol. 6, no. 2, pp. 143–155, 2009. 4, 9

[78] K. Liu and C. Xiong, "A novel 10-dof exoskeleton rehabilitation robot based on the postural synergies of upper extremity movements," in Inter-

national Conference on Intelligent Robotics and Applications. Springer, 2013, pp. 363–372. 4, 9, 13

[79] X. Jiang, C. Xiong, R. Sun, and Y. Xiong, "Characteristics of the robotic arm of a 9-dof upper limb rehabilitation robot powered by pneumatic muscles," in International Conference on Intelligent Robotics and Applications. Springer, 2010, pp. 463–474. 4, 9, 13, 16

[80] X.-Z. Jiang, C.-H. Xiong, R.-L. Sun, and Y.-L. Xiong, "Characteristics of the robotic joint of a 9-dof upper limb rehabilitation robot driven by pneumatic muscles," International Journal of Humanoid Robotics, vol. 8, no. 04, pp. 743–760, 2011. 4, 9, 13, 16

[81] T. G. Sugar, J. He, E. J. Koeneman, J. B. Koeneman, R. Herman, H. Huang, R. S. Schultz, D. Herring, J. Wanberg, S. Balasubramanian et al., "Design and control of rupert: a device for robotic upper extremity repetitive therapy," IEEE transactions on neural systems and rehabilitation engineering, vol. 15, no. 3, pp. 336–346, 2007. 4, 6, 9

[82] J. He, E. Koeneman, R. Schultz, D. Herring, J. Wanberg, H. Huang, T. Sugar, R. Herman, and J. Koeneman, "Rupert: a device for robotic upper extremity repetitive therapy," in 2005 IEEE Engineering in Medicine and Biology 27th Annual Conference. IEEE, 2006, pp. 6844–6847. 4, 6, 9

[83] J. He, E. J. Koeneman, R. Schultz, H. Huang, J. Wanberg, D. Herring, T. Sugar, R. Herman, and J. Koeneman, "Design of a robotic upper extremity repetitive therapy device," in 9th International Conference on Rehabilitation Robotics, 2005. ICORR 2005. IEEE, 2005, pp. 95–98. 4, 6, 9

[84] J. Wu, J. Huang, Y. Wang, and K. Xing, "A wearable rehabilitation robotic hand driven by pm-ts actuators," in International Conference on Intelligent Robotics and Applications. Springer, 2010, pp. 440–450. 4, 6, 8, 16

[85] T. Chen, R. Casas, and P. S. Lum, "An elbow exoskeleton for upper limb rehabilitation with series elastic actuator and cable-driven differential," IEEE Transactions on Robotics, vol. 35, no. 6, pp. 1464–1474, 2019. 4, 8, 16

[86] H.-S. Park, Y. Ren, and L.-Q. Zhang, "Intelliarm: An exoskeleton for diagnosis and treatment of patients with neurological impairments," in 2008 2nd IEEE RAS & EMBS International Conference on Biomedical Robotics and Biomechatronics. IEEE, 2008, pp. 109–114. 6, 10, 15

[87] A. Frisoli, L. Borelli, A. Montagner, S. Marcheschi, C. Procopio, F. Salsedo, M. Bergamasco, M. C. Carboncini, M. Tolaini, and B. Rossi, "Arm rehabilitation with a robotic exoskeleleton in virtual reality," in 2007 IEEE 10th International Conference on Rehabilitation Robotics. IEEE, 2007, pp. 631–642. 6, 9, 13, 16

[88] Y. Fu, Q. Zhang, F. Zhang, and Z. Gan, "Design and development of a hand rehabilitation robot for patient-cooperative therapy following stroke," in 2011 IEEE International Conference on Mechatronics and Automation. IEEE, 2011, pp. 112–117. 6, 8, 17

[89] G. Johnson, D. Carus, G. Parrini, S. Marchese, and R. Valeggi, "The design of a five-degree-of-freedom powered orthosis for the upper limb," Proceedings of the Institution of Mechanical Engineers, Part H: Journal of Engineering in Medicine, vol. 215, no. 3, pp. 275–284, 2001. 6, 9, 13, 16

[90] G. Rosati, D. Zanotto, R. Secoli, and A. Rossi, "Design and control of two planar cable-driven robots for upper-limb neurorehabilitation," in 2009 IEEE International Conference on Rehabilitation Robotics. IEEE, 2009, pp. 560–565. 6, 10, 13, 15, 16, 17

[91] G. Rosati, M. Andreolli, A. Biondi, and P. Gallina, "Performance of cable suspended robots for upper limb rehabilitation," in 2007 IEEE 10th International Conference on Rehabilitation Robotics. IEEE, 2007, pp. 385–392. 6, 10, 13, 19

[92] S. Masiero, A. Celia, G. Rosati, and M. Armani, "Robotic-assisted rehabilitation of the upper limb after acute stroke," Archives of physical medicine and rehabilitation, vol. 88, no. 2, pp. 142–149, 2007. 6

[93] S. Masiero, M. Armani, G. Ferlini, G. Rosati, and A. Rossi, "Randomized trial of a robotic assistive device for the upper extremity during early inpatient stroke rehabilitation," Neurorehabilitation and neural repair, vol. 28, no. 4, pp. 377–386, 2014. 6

[94] T. N. Bryce, M. P. Dijkers, and A. J. Kozlowski, "Framework for assessment of the usability of lower-extremity robotic exoskeletal orthoses," American journal of physical medicine & rehabilitation, vol. 94, no. 11, pp. 1000–1014, 2015. 6

[95] J. F. Veneman, R. Ekkelenkamp, R. Kruidhof, F. C. van der Helm, and H. van der Kooij, "A series elastic-and bowden-cable-based actuation system for use as torque actuator in exoskeleton-type robots," The inter-

national journal of robotics research, vol. 25, no. 3, pp. 261–281, 2006. 6, 11, 13, 16, 19

[96] J. F. Veneman, R. Kruidhof, E. E. Hekman, R. Ekkelenkamp, E. H. Van Asseldonk, and H. Van Der Kooij, "Design and evaluation of the lopes exoskeleton robot for interactive gait rehabilitation," IEEE Transactions on Neural Systems and Rehabilitation Engineering, vol. 15, no. 3, pp. 379–386, 2007. 6, 7, 11, 13, 16, 17, 19

[97] R. Ekkelenkamp, J. Veneman, and H. Van Der Kooij, "Lopes: a lower extremity powered exoskeleton," in Proceedings 2007 IEEE International Conference on Robotics and Automation. IEEE, 2007, pp. 3132–3133. 6, 11, 13, 16, 19

[98] X. Jin, X. Cui, and S. K. Agrawal, "Design of a cable-driven active leg exoskeleton (c-alex) and gait training experiments with human subjects," in 2015 IEEE International Conference on Robotics and Automation (ICRA). IEEE, 2015, pp. 5578–5583. 6, 7, 11, 15, 17, 19

[99] J. Park, S. Park, C. Kim, J. H. Park, and J. Choi, "Design and control of a powered lower limb orthosis using a cable-differential mechanism, cowalk-mobile 2," IEEE Access, vol. 9, pp. 43 775–43 784, 2021. 6

[100] A. M. Barbosa, J. C. M. Carvalho, and R. S. Gonçalves, "Cable-driven lower limb rehabilitation robot," Journal of the Brazilian Society of Mechanical Sciences and Engineering, vol. 40, no. 5, p. 245, 2018. 6, 11, 13, 16

[101] Y.-L. Park, B.-r. Chen, N. O. Pérez-Arancibia, D. Young, L. Stirling, R. J. Wood, E. C. Goldfield, and R. Nagpal, "Design and control of a bio-inspired soft wearable robotic device for ankle–foot rehabilitation," Bioinspiration & biomimetics, vol. 9, no. 1, p. 016007, 2014. 6, 11

[102] N. Dhir, H. Dallali, E. M. Ficanha, G. A. Ribeiro, and M. Rastgaar, "Locomotion envelopes for adaptive control of powered ankle prostheses," in 2018 IEEE International Conference on Robotics and Automation (ICRA). IEEE, 2018, pp. 1488–1495. 6, 11, 13, 16

[103] G. Abbasnejad, J. Yoon, and H. Lee, "Optimum kinematic design of a planar cable-driven parallel robot with wrench-closure gait trajectory," Mechanism and Machine Theory, vol. 99, pp. 1–18, 2016. 7, 11

[104] S. Gharatappeh, G. Abbasnejad, J. Yoon, and H. Lee, "Control of cable-driven parallel robot for gait rehabilitation," in 2015 12th International Conference on Ubiquitous Robots and Ambient Intelligence (URAI). IEEE, 2015, pp. 377–381. 7, 11

[105] A. Taherifar, M. Hadian, M. Mousavi, A. Rassaf, and F. Ghiasi, "Lokoiran-a novel robot for rehabilitation of spinal cord injury and stroke patients," in 2013 First RSI/ISM International Conference on Robotics and Mechatronics (ICRoM). IEEE, 2013, pp. 218–223. 7, 11, 15, 17

[106] D. Surdilovic and R. Bernhardt, "String-man: a new wire robot for gait rehabilitation," in IEEE International Conference on Robotics and Automation, 2004. Proceedings. ICRA'04. 2004, vol. 2. IEEE, 2004, pp. 2031–2036. 7, 11, 13, 19

[107] B. M. Fleerkotte, J. H. Buurke, B. Koopman, L. Schaake, H. van der Kooij, E. H. van Asseldonk, and J. S. Rietman, "Effectiveness of the lower extremity powered exoskeleton (lopes) robotic gait trainer on ability and quality of walking in sci patients," in Converging Clinical and Engineering Research on Neurorehabilitation. Springer, 2013, pp. 161–165. 7

[108] J. Bae, C. Siviy, M. Rouleau, N. Menard, K. O'Donnell, I. Geliana, M. Athanassiu, D. Ryan, C. Bibeau, L. Sloot et al., "A lightweight and efficient portable soft exosuit for paretic ankle assistance in walking after stroke," in 2018 IEEE International Conference on Robotics and Automation (ICRA). IEEE, 2018, pp. 2820–2827. 7, 11, 18, 19

[109] M. Wu, T. G. Hornby, J. M. Landry, H. Roth, and B. D. Schmit, "A cable-driven locomotor training system for restoration of gait in human sci," Gait & posture, vol. 33, no. 2, pp. 256–260, 2011. 7, 11

[110] J. Kang, V. Vashista, and S. K. Agrawal, "On the adaptation of pelvic motion by applying 3-dimensional guidance forces using tpad," IEEE Transactions on Neural Systems and Rehabilitation Engineering, vol. 25, no. 9, pp. 1558–1567, 2017. 7, 11, 15, 17

[111] B. Zi, G. Yin, and D. Zhang, "Design and optimization of a hybrid-driven waist rehabilitation robot," Sensors, vol. 16, no. 12, p. 2121, 2016. 7, 10

[112] B. Zi, G. Yin, Y. Li, and D. Zhang, "Kinematic performance analysis of a hybrid-driven waist rehabilitation robot," in Mechatronics and Robotics Engineering for Advanced and Intelligent Manufacturing. Springer, 2017, pp. 73–86. 7

[113] Q. Chen, B. Zi, Z. Sun, Y. Li, and Q. Xu, "Design and development of a new cable-driven parallel robot for waist rehabilitation," IEEE/ASME Transactions on Mechatronics, vol. 24, no. 4, pp. 1497–1507, 2019. 7, 10, 13, 16

[114] T. Zhao, S. Qian, Q. Chen, and Z. Sun, "Design and analysis of a cable-driven parallel robot for waist rehabilitation," in 2018 IEEE International Conference on Mechatronics, Robotics and Automation (ICMRA). IEEE, 2018, pp. 173–178. 7, 10

[115] Z. Yang, B. Zi, and B. Chen, "Mechanism design and kinematic analysis of a waist and lower limbs cable-driven parallel rehabilitation robot," in 2019 IEEE 3rd Advanced Information Management, Communicates, Electronic and Automation Control Conference (IMCEC). IEEE, 2019, pp. 723–727. 7, 10

[116] Z.-F. Shao, X. Tang, and W. Yi, "Optimal design of a 3-dof cable-driven upper arm exoskeleton," Advances in Mechanical Engineering, vol. 6, p. 157096, 2014. 8

[117] W. Wei, Z. Qu, W. Wang, P. Zhang, and F. Hao, "Design on the bowden cable-driven upper limb soft exoskeleton," Applied bionics and biomechanics, vol. 2018, 2018. 8, 15

[118] A. V. Dowling, O. Barzilay, Y. Lombrozo, and A. Wolf, "An adaptive home-use robotic rehabilitation system for the upper body," IEEE journal of translational engineering in health and medicine, vol. 2, pp. 1–10, 2014. 8

[119] G. Zuccon, M. Bottin, M. Ceccarelli, and G. Rosati, "Design and performance of an elbow assisting mechanism," Machines, vol. 8, no. 4, p. 68, 2020. 8

[120] M. Ceccarelli, L. Ferrara, and V. Petuya, "Design of a cable-driven device for elbow rehabilitation and exercise," in Interdisciplinary Applications of Kinematics. Springer, 2019, pp. 61–68. 8

[121] D. H. Kim and H.-S. Park, "Cable actuated dexterous (cadex) glove for effective rehabilitation of the hand for patients with neurological diseases," in 2018 IEEE/RSJ International Conference on Intelligent Robots and Systems (IROS). IEEE, 2018, pp. 2305–2310. 8

[122] S. Mohamaddan and T. Komeda, "Wire-driven mechanism for finger rehabilitation device," in 2010 IEEE International Conference on Mechatronics and Automation. IEEE, 2010, pp. 1015–1018. 8

[123] H. In, K.-J. Cho, K. Kim, and B. Lee, "Jointless structure and underactuation mechanism for compact hand exoskeleton," in 2011 IEEE International Conference on Rehabilitation Robotics. IEEE, 2011, pp. 1–6. 8

[124] U. Jeong, H.-K. In, and K.-J. Cho, "Implementation of various control algorithms for hand rehabilitation exercise using wearable robotic hand," Intelligent Service Robotics, vol. 6, no. 4, pp. 181–189, 2013. 8, 16

[125] C. L. Jones, F. Wang, R. Morrison, N. Sarkar, and D. G. Kamper, "Design and development of the cable actuated finger exoskeleton for hand rehabilitation following stroke," IEEE/ASME Transactions on Mechatronics, vol. 19, no. 1, pp. 131–140, 2012. 8, 13

[126] M. Cempini, S. M. M. De Rossi, T. Lenzi, M. Cortese, F. Giovacchini, N. Vitiello, and M. C. Carrozza, "Kinematics and design of a portable and wearable exoskeleton for hand rehabilitation," in 2013 IEEE 13th International Conference on Rehabilitation Robotics (ICORR). IEEE, 2013, pp. 1–6. 8

[127] S. Wang, J. Li, and R. Zheng, "A resistance compensation control algorithm for a cable-driven hand exoskeleton for motor function rehabilitation," in International Conference on Intelligent Robotics and Applications. Springer, 2010, pp. 398–404. 8, 15

[128] S. Liu, D. Meng, L. Cheng, and M. Chen, "An iterative learning controller for a cable-driven hand rehabilitation robot," in IECON 2017-43rd Annual Conference of the IEEE Industrial Electronics Society. IEEE, 2017, pp. 5701–5706. 8, 15

[129] G. N. DeSouza, P. Aubin, K. Petersen, H. Sallum, C. Walsh, A. Correia, and L. Stirling, "A pediatric robotic thumb exoskeleton for at-home rehabilitation," International Journal of Intelligent Computing and Cybernetics, 2014. 8

[130] J. Ma, R. Mo, M. Chen, L. Cheng, and H. Qi, "Mirror-training of a cable-driven hand rehabilitation robot based on surface electromyography (semg)," in 2019 Tenth International Conference on Intelligent Control and Information Processing (ICICIP). IEEE, 2019, pp. 309–315. 8

[131] S. Yu, H. Perez, J. Barkas, M. Mohamed, M. Eldaly, T.-H. Huang, X. Yang, H. Su, M. del Mar Cortes, and D. J. Edwards, "A soft high force hand exoskeleton for rehabilitation and assistance of spinal cord injury and stroke individuals," in Frontiers in Biomedical Devices, vol. 41037. American Society of Mechanical Engineers, 2019, p. V001T09A011. 8

[132] M. M. Ullah, U. Hafeez, M. N. Shehzad, M. N. Awais, and H. Elahi, "A soft robotic glove for assistance and rehabilitation of stroke affected patients," in 2019 International Conference on Frontiers of Information Technology (FIT). IEEE, 2019, pp. 110–1105. 8

[133] M. S. John, N. Thomas, and V. Sivakumar, "Design and development of cable driven upper limb exoskeleton for arm rehabilitation," Int. J. Sci. Eng. Res, vol. 7, no. 3, pp. 1432–1440, 2016. 9, 17

[134] N. Jarrassé, J. Robertson, P. Garrec, J. Paik, V. Pasqui, Y. Perrot, A. Roby-Brami, D. Wang, and G. Morel, "Design and acceptability assessment of a new reversible orthosis," in 2008 IEEE/RSJ International Conference on Intelligent Robots and Systems. IEEE, 2008, pp. 1933–1939. 9, 13, 16

[135] Z. Li, W. Chen, J. Zhang, and S. Bai, "Design and control of a 4-dof cable-driven arm rehabilitation robot (carr-4)," in 2017 IEEE International Conference on Cybernetics and Intelligent Systems (CIS) and IEEE Conference on Robotics, Automation and Mechatronics (RAM). IEEE, 2017, pp. 581–586. 9, 15, 17

[136] I. Gaponov, D. Popov, S. J. Lee, and J.-H. Ryu, "Auxilio: a portable cable-driven exosuit for upper extremity assistance," International Journal of Control, Automation and Systems, vol. 15, no. 1, pp. 73–84, 2017. 9

[137] W. M. Nunes, L. A. O. Rodrigues, L. P. Oliveira, J. F. Ribeiro, J. C. M. Carvalho, and R. S. Gonçalves, "Cable-based parallel manipulator for rehabilitation of shoulder and elbow movements," in 2011 IEEE international conference on rehabilitation robotics. IEEE, 2011, pp. 1–6. 9

[138] K. Shi, A. Song, Y. Li, D. Chen, and H. Li, "Cable-driven 4-dof upper limb rehabilitation robot," in 2019 IEEE/RSJ International Conference on Intelligent Robots and Systems (IROS). IEEE, 2019, pp. 6465–6472. 9, 16

[139] Z. Li, W. Li, W.-H. Chen, J. Zhang, J. Wang, Z. Fang, and G. Yang, "Mechatronics design and testing of a cable-driven upper limb rehabilitation exoskeleton with variable stiffness," Review of Scientific Instruments, vol. 92, no. 2, p. 024101, 2021. 9

[140] E. Pezent, C. G. Rose, A. D. Deshpande, and M. K. O'Malley, "Design and characterization of the openwrist: A robotic wrist exoskeleton for coordinated hand-wrist rehabilitation," in 2017 International Conference on Rehabilitation Robotics (ICORR). IEEE, 2017, pp. 720–725. 9, 13

[141] A. Schiele and G. Hirzinger, "A new generation of ergonomic exoskeletons-the high-performance x-arm-2 for space robotics telepresence," in 2011 IEEE/RSJ International Conference on Intelligent Robots and Systems. IEEE, 2011, pp. 2158–2165. 9, 13, 16

[142] F. Xiao, Y. Gao, Y. Wang, Y. Zhu, and J. Zhao, "Design and evaluation of a 7-dof cable-driven upper limb exoskeleton," Journal of Mechanical Science and Technology, vol. 32, no. 2, pp. 855–864, 2018. 9

[143] J. C. Perry, J. Rosen, and S. Burns, "Upper-limb powered exoskeleton design," IEEE/ASME transactions on mechatronics, vol. 12, no. 4, pp. 408–417, 2007. 9, 16

[144] C.-T. Chen, W.-Y. Lien, C.-T. Chen, M.-J. Twu, and Y.-C. Wu, "Dynamic modeling and motion control of a cable-driven robotic exoskeleton with pneumatic artificial muscle actuators," IEEE Access, vol. 8, pp. 149 796–149 807, 2020. 9, 15, 17

[145] J. Wang, W. Li, W. Chen, and J. Zhang, "Motion control of a 4-dof cable-driven upper limb exoskeleton," in 2019 14th IEEE Conference on Industrial Electronics and Applications (ICIEA). IEEE, 2019, pp. 2129–2134. 9

[146] N. G. Tsagarakis and D. G. Caldwell, "Development and control of a 'soft-actuated'exoskeleton for use in physiotherapy and training," Autonomous Robots, vol. 15, no. 1, pp. 21–33, 2003. 10, 16

[147] J. C. Perry and J. Rosen, "Design of a 7 degree-of-freedom upper-limb powered exoskeleton," in The First IEEE/RAS-EMBS International Conference on Biomedical Robotics and Biomechatronics, 2006. BioRob 2006. IEEE, 2006, pp. 805–810. 10, 16

[148] M. Morris and O. Masory, "Planar cable-driven rehabilitation robot," in Proceedings of the 2008 Florida Conference on Recent Advances in Robotics, 2008, pp. 8–9. 10

[149] J. Niu, Q. Yang, X. Wang, and R. Song, "Sliding mode tracking control of a wire-driven upper-limb rehabilitation robot with nonlinear disturbance observer," Frontiers in neurology, vol. 8, p. 646, 2017. 10, 15

[150] J. Niu, Q. Yang, G. Chen, and R. Song, "Nonlinear disturbance observer based sliding mode control of a cable-driven rehabilitation robot," in 2017 International Conference on Rehabilitation Robotics (ICORR). IEEE, 2017, pp. 664–669. 10, 15

[151] C. Fanin, P. Gallina, A. Rossi, U. Zanatta, and S. Masiero, "Nerebot: a wire-based robot for neurorehabilitation," in ICORR'03. HWRS-ERC, 2003, pp. 23–27. 10, 13, 19

[152] M. Stefano, P. Patrizia, A. Mario, G. Ferlini, R. Rizzello, and G. Rosati, "Robotic upper limb rehabilitation after acute stroke by nerebot: Evaluation of treatment costs," BioMed research international, vol. 2014, 2014. 10, 19

[153] G. Rosati, P. Gallina, A. Rossi, and S. Masiero, "Wire-based robots for upper-limb rehabilitation," International Journal of Assistive Robotics and Mechatronics, vol. 7, pp. 3–10, 2006. 10
[154] D. Mayhew, B. Bachrach, W. Z. Rymer, and R. F. Beer, "Development of the macarm-a novel cable robot for upper limb neurorehabilitation," in 9th International Conference on Rehabilitation Robotics, 2005. ICORR 2005. IEEE, 2005, pp. 299–302. 10
[155] X. Li, Q. Yang, and R. Song, "Performance-based hybrid control of a cable-driven upper-limb rehabilitation robot," IEEE Transactions on Biomedical Engineering, 2020. 10, 15, 17
[156] S. Kyeong, Y. Na, and J. Kim, "A mechatronic mirror-image motion device for symmetric upper-limb rehabilitation," International Journal of Precision Engineering and Manufacturing, pp. 1–10, 2020. 10, 16
[157] J. Kim, R. Heimgartner, G. Lee, N. Karavas, D. Perry, D. L. Ryan, A. Eckert-Erdheim, P. Murphy, D. K. Choe, I. Galiana et al., "Autonomous and portable soft exosuit for hip extension assistance with online walking and running detection algorithm," in 2018 IEEE International Conference on Robotics and Automation (ICRA). IEEE, 2018, pp. 1–8. 11, 17
[158] E. J. Park, J. Kang, H. Su, P. Stegall, D. L. Miranda, W.-H. Hsu, M. Karabas, N. Phipps, S. K. Agrawal, E. C. Goldfield et al., "Design and preliminary evaluation of a multi-robotic system with pelvic and hip assistance for pediatric gait rehabilitation," in 2017 International Conference on Rehabilitation Robotics (ICORR). IEEE, 2017, pp. 332–339. 11, 13, 16
[159] Y.-L. Park, J. Santos, K. G. Galloway, E. C. Goldfield, and R. J. Wood, "A soft wearable robotic device for active knee motions using flat pneumatic artificial muscles," in 2014 IEEE International Conference on Robotics and Automation (ICRA). IEEE, 2014, pp. 4805–4810. 11
[160] P. K. Jamwal, S. Xie, and K. C. Aw, "Kinematic design optimization of a parallel ankle rehabilitation robot using modified genetic algorithm," Robotics and Autonomous Systems, vol. 57, no. 10, pp. 1018–1027, 2009. 11
[161] T. Noda, A. Takai, T. Teramae, E. Hirookai, K. Hase, and J. Morimoto, "Robotizing double-bar ankle-foot orthosis," in 2018 IEEE International Conference on Robotics and Automation (ICRA). IEEE, 2018, pp. 2782–2787. 11
[162] M. Russo and M. Ceccarelli, "Analysis of a wearable robotic system for ankle rehabilitation," Machines, vol. 8, no. 3, p. 48, 2020. 11
[163] T.-m. Wang, X. Pei, T.-g. Hou, Y.-b. Fan, X. Yang, H. M. Herr, and X.-b. Yang, "An untethered cable-driven ankle exoskeleton with plantarflexion-dorsiflexion bidirectional movement assistance," Frontiers of Information Technology & Electronic Engineering, vol. 21, pp. 723–739, 2020. 11
[164] T. Lee, I. Kim, and Y. S. Baek, "Design of a 2dof ankle exoskeleton with a polycentric structure and a bi-directional tendon-driven actuator controlled using a pid neural network," in Actuators, vol. 10, no. 1. Multidisciplinary Digital Publishing Institute, 2021, p. 9. 11, 16
[165] S. Lee, N. Karavas, B. T. Quinlivan, D. LouiseRyan, D. Perry, A. Eckert-Erdheim, P. Murphy, T. G. Goldy, N. Menard, M. Athanassiu et al., "Autonomous multi-joint soft exosuit for assistance with walking overground," in 2018 IEEE International Conference on Robotics and Automation (ICRA). IEEE, 2018, pp. 2812–2819. 11, 15
[166] N. Costa, M. Bezdicek, M. Brown, J. O. Gray, D. G. Caldwell, and S. Hutchins, "Joint motion control of a powered lower limb orthosis for rehabilitation," International Journal of Automation and Computing, vol. 3, no. 3, pp. 271–281, 2006. 11, 13, 16
[167] S.-H. Hyon, J. Morimoto, T. Matsubara, T. Noda, and M. Kawato, "Xor: Hybrid drive exoskeleton robot that can balance," in 2011 IEEE/RSJ International Conference on Intelligent Robots and Systems. IEEE, 2011, pp. 3975–3981. 11, 13, 16
[168] D. P. Ferris and C. L. Lewis, "Robotic lower limb exoskeletons using proportional myoelectric control," in 2009 Annual International Conference of the IEEE Engineering in Medicine and Biology Society. IEEE, 2009, pp. 2119–2124. 11
[169] M. Wehner, B. Quinlivan, P. M. Aubin, E. Martinez-Villalpando, M. Baumann, L. Stirling, K. Holt, R. Wood, and C. Walsh, "A lightweight soft exosuit for gait assistance," in 2013 IEEE international conference on robotics and automation. IEEE, 2013, pp. 3362–3369. 11
[170] H. Faqihi, M. Saad, K. Benjelloun, M. Benbrahim, and M. N. Kabbaj, "Tracking trajectory of a cable-driven robot for lower limb rehabilitation," International Journal of Electrical, Computer, Energetic, Electronic and Communication Engineering, vol. 10, no. 8, pp. 1036–1041, 2016. 11
[171] Y. Liu, Y. Gao, F. Xiao, and J. Zhao, "Research on the cable-pulley underactuated lower limb exoskeleton," in 2017 IEEE International Conference on Mechatronics and Automation (ICMA). IEEE, 2017, pp. 577–583. 11, 15, 17
[172] A. Alamdari and V. Krovi, "Design and analysis of a cable-driven articulated rehabilitation system for gait training," Journal of Mechanisms and Robotics, vol. 8, no. 5, p. 051018, 2016. 11
[173] A. Alamdari, R. Haghighi, and V. Krovi, "Gravity-balancing of elastic articulated-cable leg-orthosis emulator," Mechanism and machine theory, vol. 131, pp. 351–370, 2019. 11
[174] A. Alamdari, "Cable-driven articulated rehabilitation system for gait training," Ph.D. dissertation, State University of New York at Buffalo, 2016. 11
[175] K. Homma, O. Fukuda, and Y. Nagata, "Study of a wire-driven leg rehabilitation system," in IEEE/RSJ international conference on intelligent robots and systems, vol. 2. IEEE, 2002, pp. 1451–1456. 11
[176] K. Homma, O. Fukuda, J. Sugawara, Y. Nagata, and M. Usuba, "A wire-driven leg rehabilitation system: Development of a 4-dof experimental system," in Proceedings 2003 IEEE/ASME International Conference on Advanced Intelligent Mechatronics (AIM 2003), vol. 2. IEEE, 2003, pp. 908–913. 11
[177] D. Cafolla, M. Russo, and G. Carbone, "Cube, a cable-driven device for limb rehabilitation," Journal of Bionic Engineering, vol. 16, no. 3, pp. 492–502, 2019. 11
[178] ——, "Design of cube, a cable-driven device for upper and lower limb exercising," in New Trends in Medical and Service Robotics. Springer, 2019, pp. 255–263. 11
[179] S. S. Iyer, J. V. Joseph, N. Sanjeevi, Y. Singh, and V. Vashista, "Development and applicability of a cable-driven wearable adaptive rehabilitation suit (wears)," in 2019 28th IEEE International Conference on Robot and Human Interactive Communication (RO-MAN). IEEE, 2019, pp. 1–6. 11
[180] Z. Zhou, Z. Wang, and Q. Wang, "On the design of rigid-soft hybrid exoskeleton based on remote cable actuator for gait rehabilitation," in 2020 IEEE/ASME International Conference on Advanced Intelligent Mechatronics (AIM). IEEE, 2020, pp. 1902–1907. 11
[181] Y.-J. Kim, C.-K. Park, and K. G. Kim, "An emg-based variable impedance control for elbow exercise: preliminary study," Advanced Robotics, vol. 31, no. 15, pp. 809–820, 2017. 13
[182] N. Nazmi, M. A. A. Rahman, S. A. Mazlan, H. Zamzuri, and M. Mizukawa, "Electromyography (emg) based signal analysis for physiological device application in lower limb rehabilitation," in 2015 2nd International Conference on Biomedical Engineering (ICoBE). IEEE, 2015, pp. 1–6. 13
[183] D. Czarkowski and T. O'Mahony, "Intelligent controller design based on gain and phase margin specifications," 2004. 13
[184] N. Hogan, "Impedance control: An approach to manipulation: Part i—theory," 1985. 13
[185] P. Lammertse, "Admittance control and impedance control-a dual," FCS Control Systems, 2004. 13
[186] E. Arabi, B. C. Gruenwald, T. Yucelen, and J. E. Steck, "Guaranteed model reference adaptive control performance in the presence of actuator failures," in AIAA Information Systems-AIAA Infotech@ Aerospace, 2017, p. 0669. 15
[187] Y. Mao, X. Jin, and S. K. Agrawal, "Real-time estimation of glenohumeral joint rotation center with cable-driven arm exoskeleton (carex)—a cable-based arm exoskeleton," Journal of mechanisms and robotics, vol. 6, no. 1, 2014. 15
[188] N. Hogan, "Impedance control: An approach to manipulation," in 1984 American control conference. IEEE, 1984, pp. 304–313. 17
[189] R. Riener, L. Lunenburger, S. Jezernik, M. Anderschitz, G. Colombo, and V. Dietz, "Patient-cooperative strategies for robot-aided treadmill training: first experimental results," IEEE transactions on neural systems and rehabilitation engineering, vol. 13, no. 3, pp. 380–394, 2005. 17, 18
[190] J. C. P. Ibarra and A. A. Siqueira, "Impedance control of rehabilitation robots for lower limbs, review," in 2014 Joint Conference on Robotics: SBR-LARS Robotics Symposium and Robocontrol. IEEE, 2014, pp. 235–240. 17
[191] Y. H. Tsoi and S. Q. Xie, "Impedance control of ankle rehabilitation robot," in 2008 IEEE International Conference on Robotics and Biomimetics. IEEE, 2009, pp. 840–845. 17
[192] S. Chiaverini, B. Siciliano, and L. Villani, "A survey of robot interaction control schemes with experimental comparison," IEEE/ASME Transactions on mechatronics, vol. 4, no. 3, pp. 273–285, 1999. 17

[193] D. E. Whitney, "Historical perspective and state of the art in robot force control," The International Journal of Robotics Research, vol. 6, no. 1, pp. 3–14, 1987. 17

[194] C. Ott, R. Mukherjee, and Y. Nakamura, "Unified impedance and admittance control," in 2010 IEEE International Conference on Robotics and Automation. IEEE, 2010, pp. 554–561. 18

[195] B. K. Dinh, M. Xiloyannis, L. Cappello, C. W. Antuvan, S.-C. Yen, and L. Masia, "Adaptive backlash compensation in upper limb soft wearable exoskeletons," Robotics and Autonomous Systems, vol. 92, pp. 173–186, 2017. 18

[196] A. U. Pehlivan, F. Sergi, and M. K. O'Malley, "Adaptive control of a serial-in-parallel robotic rehabilitation device," in 2013 IEEE 13th International Conference on Rehabilitation Robotics (ICORR). IEEE, 2013, pp. 1–6. 18

[197] P. E. Crago, N. Lan, P. H. Veltink, J. J. Abbas, and C. Kantor, "New control strategies for neuroprosthetic systems," Journal of rehabilitation research and development, vol. 33, pp. 158–172, 1996. 18

[198] Q.-T. Dao and S.-i. Yamamoto, "Assist-as-needed control of a robotic orthosis actuated by pneumatic artificial muscle for gait rehabilitation," Applied Sciences, vol. 8, no. 4, p. 499, 2018. 18

[199] S. Younis and N. Z. Azlan, "Assist-as-needed control strategies for upper limb rehabilitation therapy: A review," Jurnal Mekanikal, vol. 42, no. 1, 2019. 18

[200] S. Y. A. Mounis, N. Z. Azlan, and F. Sado, "Assist-as-needed control strategy for upper-limb rehabilitation based on subject's functional ability," Measurement and Control, vol. 52, no. 9-10, pp. 1354–1361, 2019. 18

[201] B. Chen, H. Ma, L.-Y. Qin, F. Gao, K.-M. Chan, S.-W. Law, L. Qin, and W.-H. Liao, "Recent developments and challenges of lower extremity exoskeletons," Journal of Orthopaedic Translation, vol. 5, pp. 26–37, 2016. 18

[202] J. Stein, H. I. Krebs, W. R. Frontera, S. E. Fasoli, R. Hughes, and N. Hogan, "Comparison of two techniques of robot-aided upper limb exercise training after stroke," American journal of physical medicine & rehabilitation, vol. 83, no. 9, pp. 720–728, 2004. 18

[203] L. L. Cai, A. J. Fong, C. K. Otoshi, Y. Liang, J. W. Burdick, R. R. Roy, and V. R. Edgerton, "Implications of assist-as-needed robotic step training after a complete spinal cord injury on intrinsic strategies of motor learning," Journal of Neuroscience, vol. 26, no. 41, pp. 10 564–10 568, 2006. 18

[204] Y. Cho, B. Kang, C. Park, and J. Cheong, "Kinematics of elastic tendons for tendon-driven manipulators with transmission friction," IEEE/ASME Transactions on Mechatronics, 2021. 20

[205] A. Tonazzini, S. Mintchev, B. Schubert, B. Mazzolai, J. Shintake, and D. Floreano, "Variable stiffness fiber with self-healing capability," Advanced Materials, vol. 28, no. 46, pp. 10 142–10 148, 2016. 20

[206] Y.-J. Kim, S. Cheng, S. Kim, and K. Iagnemma, "Design of a tubular snake-like manipulator with stiffening capability by layer jamming," in 2012 IEEE/RSJ International Conference on Intelligent Robots and Systems. IEEE, 2012, pp. 4251–4256. 20

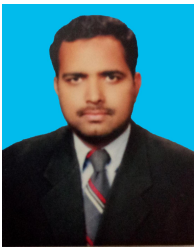

MUHAMMAD SHOAIB received the master's degree from the Department of Control and Instrumentation Engineering, Korea University, Sejong, South Korea, in 2017. He worked as research assistant in the Research Laboratory for Advanced Robotics at Korea University. He is currently pursuing the Ph.D. degree from RMIT University, Melbourne, Australia. His research interests include cable-driven parallel mechanisms, cable-driven rehabilitation manipulators, mobile manipulators, robotic hands and grasping, and sensor based mechanical system controls.

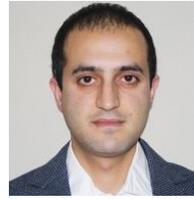

EHSAN ASADI received the B.Sc. and M.Eng. degrees in mechanical engineering and the Ph.D. degree from Politecnico di Milano, Milan, Italy. He is currently a Senior Lecturer within the Discipline of Manufacturing, Materials and Mechatronics in the School of Engineering, RMIT University. Prior to this, he worked as postdoctoral research fellow in the Robotic Research Center at Nanyang Technological University, Singapore, and co-founded a robotic startup company, Transforma Robotics Pte. Ltd.. He continues his research in mechatronics, sensor fusion, robotic vision, and intelligent robotics for filed applications.

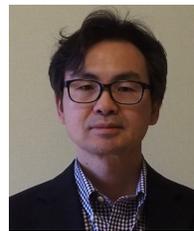

JOONO CHEONG received BS, MS and PhD all from Pohang University of Science and Technology (POSTECH) in 1995, 1997 and 2003, respectively. In 2003, he was a Researcher with the Institute of Precision Machine and Design, Seoul National University, Seoul. From 2003 to 2005, he was a Postdoc Researcher of the Research Laboratory of Electronics at Massachusetts Institute of Technology, Cambridge, MA. Since 2005, he has been with the Department of Control and Instrumentation Engineering, Korea University, Sejong, where he is currently a Professor. He is the Director of the Laboratory for Advanced Robotics at Korea University. His research interests are robotic manipulation, grasping, and mechanical systems control.

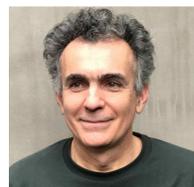

ALIREZA BAB-HADIASHAR received the B.Sc. and M.Eng. degrees in mechanical engineering and the Ph.D. degree in robotics from Monash University. He has held various positions in Monash University, the Swinburne University of Technology, and RMIT University. He is currently a Professor of mechatronics with RMIT University, where he leads the Intelligent Automation Research Group. He has published numerous highly cited articles in intelligent robotics and machine vision. His main research interests include mechatronics, intelligent automation, and robotic vision.

. . .